\documentclass[10pt,twocolumn,letterpaper]{article}

\usepackage{cvpr}              

\usepackage[table, svgnames, dvipsnames]{xcolor} 

\usepackage{array,multirow} 
\usepackage{siunitx} 
\usepackage[normalem]{ulem} 
\usepackage{adjustbox} 
\usepackage{algorithm}      
\usepackage{algorithmic}    
\usepackage{pifont} 

\def\B{\bf}
\robustify\B
\def\U{\uline}
\robustify\U

\newcolumntype{L}[1]{>{\raggedright\let\newline\\\arraybackslash\hspace{0pt}}m{#1}} 
\newcolumntype{R}[1]{>{\raggedleft\let\newline\\\arraybackslash\hspace{0pt}}m{#1}}
\newcolumntype{C}[1]{>{\centering\let\newline\\\arraybackslash\hspace{0pt}}m{#1}}

\newcommand{\cmark}{\ding{51}}%

\newcommand{\circnum}[1]{\scalebox{1.1}{\textcircled{\raisebox{0.1pt}{\scalebox{0.65}{#1}}}}}%

\usepackage[accsupp]{./axessibility} 

\definecolor{cvprblue}{rgb}{0.21,0.49,0.74}
\usepackage[pagebackref,breaklinks,colorlinks,citecolor=cvprblue]{hyperref}

\def\paperID{11438} 
\def\confName{CVPR}
\def\confYear{2024}

\title{AZ-NAS: Assembling Zero-Cost Proxies for Network Architecture Search}

\author{
  Junghyup Lee$^{1}$ \quad\quad\quad Bumsub Ham$^{1,2}$\thanks{Corresponding author.} \vspace*{0.2cm}\\
  $^{1}$Yonsei University \hspace{8mm} $^{2}$Korea Institute of Science and Technology~(KIST)\vspace*{0.2cm} \\ 
  {\url{https://cvlab.yonsei.ac.kr/projects/AZNAS}}
}

\begin{document}
\maketitle
\begin{abstract}
Training-free network architecture search~(NAS) aims to discover high-performing networks with zero-cost proxies, capturing network characteristics related to the final performance. However, network rankings estimated by previous training-free NAS methods have shown weak correlations with the performance. To address this issue, we propose AZ-NAS, a novel approach that leverages the ensemble of various zero-cost proxies to enhance the correlation between a predicted ranking of networks and the ground truth substantially in terms of the performance. To achieve this, we introduce four novel zero-cost proxies that are complementary to each other, analyzing distinct traits of architectures in the views of expressivity, progressivity, trainability, and complexity. The proxy scores can be obtained simultaneously within a single forward and backward pass, making an overall NAS process highly efficient. In order to integrate the rankings predicted by our proxies effectively, we introduce a non-linear ranking aggregation method that highlights the networks highly-ranked consistently across all the proxies. Experimental results conclusively demonstrate the efficacy and efficiency of AZ-NAS, outperforming state-of-the-art methods on standard benchmarks, all while maintaining a reasonable runtime cost.
\end{abstract}

 \section{Introduction} \label{sec:intro}
Representative neural network architectures, including ResNets~\cite{he2016deep}, MobileNets~\cite{howard2017mobilenets,sandler2018mobilenetv2}, and Vision Transformers~\cite{dosovitskiy2020image,liu2021swin}, have been developed by experts through tedious trial-and-error processes, making it hard to design new architectures for various configurations. To overcome this problem, network architecture search~(NAS) has emerged as a promising paradigm, capable of automatically seeking top-performing architectures under specified constraints. Recently, training-free NAS~\cite{mellor2021neural,chen2020tenas,abdelfattah2021zerocost,lin2021zen,li2023zico} has gained significant attention primarily due to its remarkable efficiency. It reduces computational and time costs of a NAS process drastically in comparison to earlier methods using an iterative training~\cite{baker2017designing,real2019regularized} or training parameter-shared networks~\cite{pham2018efficient,liu2019darts,cai2020once,zhao2021few}.

\begin{figure}[t]
  \small
  \begin{center}
     \includegraphics[width=0.95\columnwidth]{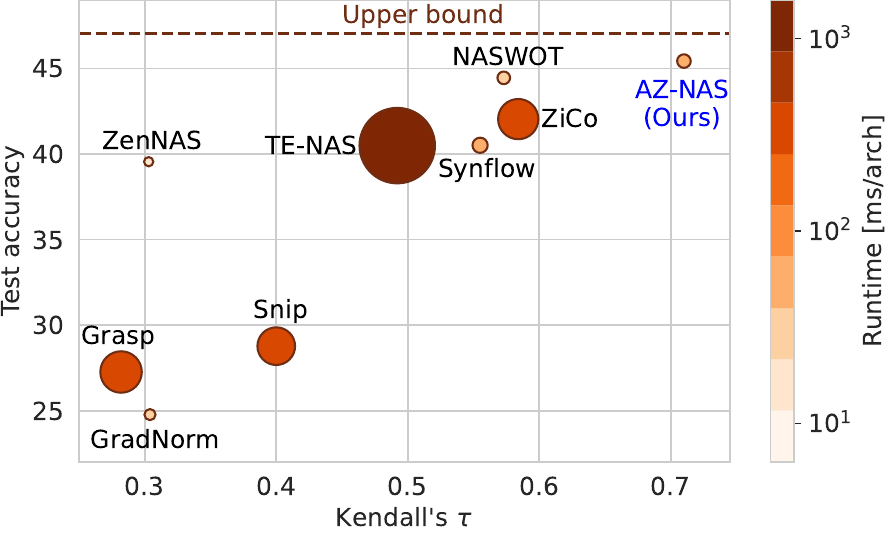}
  \end{center}
    \vspace{-0.5cm}
     \caption{Comparison of training-free NAS methods on ImageNet16-120 of NAS-Bench-201~\cite{dong2020nasbench201}. We compare correlation coefficients~(Kendall's $\tau$) between predicted rankings of networks and the ground truth in the x-axis, and test accuracies for the selected networks in the y-axis. The runtime costs are visualized by the circle size and color. By assembling the proposed zero-cost proxies, AZ-NAS achieves the best consistency between predicted rankings of the networks and the ground truth efficiently, which helps to find the network with the highest accuracy.}
    \vspace{-0.6cm}
  \label{fig:teaser}
\end{figure}

The training-free NAS methods predict the ranking of candidate networks in terms of performance via zero-cost proxies designed by empirical insights or theoretical evidences. These proxies analyze activations or gradients to capture \eg, an ability to dissect an input space into linear regions~\cite{chen2020tenas,mellor2021neural}, saliency of weights/channels~\cite{abdelfattah2021zerocost,lee2019snip,wang2020picking,tanaka2020pruning}, or training dynamics~\cite{chen2020tenas,shu2021nasi}. However, they are often less practical than the training-based NAS methods mainly due to the weak correlations with the final performance of networks. For example, recent studies~\cite{ning2021evaluating,li2023zico} point out that the simple proxies based on the number of parameters~(\#Params) or floating-point operations~(FLOPs) often provide better or competitive ranking consistency w.r.t the final performance of networks on the NAS benchmark~\cite{dong2020nasbench201}, compared to many training-free NAS methods~\cite{abdelfattah2021zerocost,wang2020picking,lee2019snip,tanaka2020pruning,chen2020tenas,mellor2021neural,lin2021zen}. This limitation might arise from the fact that the current training-free NAS methods evaluate networks from narrow perspectives, typically relying on a single proxy only~\cite{abdelfattah2021zerocost,mellor2021neural,lin2021zen,zhang2022gradsign,li2023zico}. Considering a single network characteristic might not suffice to accurately predict the network ranking without training, since various factors can significantly influence the final performance. For instance, networks with a large number of parameters do not always achieve better performance due to~\eg, gradient vanishing or exploding problems, which could not be identified if we exploit \#Params solely as a proxy.

In this paper, we introduce a novel training-free NAS method, AZ-NAS, that assembles multiple zero-cost proxies to evaluate networks from various perspectives. Naively assembling existing zero-cost proxies is, however, less effective due to the following reasons. First, it has been shown that gradient-based proxies~\cite{abdelfattah2021zerocost,lee2019snip,wang2020picking,shu2021nasi} are correlated with each other theoretically, providing similar NAS results~\cite{shu2022unifying}. This suggests that directly using an ensemble of them could prevent a comprehensive evaluation of networks, offering a limited improvement~\cite{ning2021evaluating} while requiring additional computational costs. Second, several zero-cost proxies~\cite{chen2020tenas,zhang2022gradsign} suffer from computational inefficiency, impeding scalability to large search spaces especially when they are combined with other proxies. To address these problems, we devise novel zero-cost proxies that can be obtained efficiently and capture unique network characteristics complementing each other, where the proxy scores show positive correlations with the network's performance. Each of them leverages activations, gradients, or architectural information~(\ie, FLOPs) to assess networks comprehensively in terms of expressivity, progressivity, trainability, and complexity. The proxy scores are computed simultaneously within a single forward and backward pass, making the entire NAS process efficient. We also present a non-linear ranking aggregation method to effectively merge the rankings provided by the proxies, preferring highly-ranked networks across all the proxies. We show that AZ-NAS achieves the best ranking consistency w.r.t the performance on NAS-Bench-201~\cite{dong2020nasbench201}, and it finds networks showing the state-of-the-art performance on the large-scale search spaces~\cite{sandler2018mobilenetv2,lin2021zen,chen2021autoformer}, with a reasonable runtime cost~(Fig.~\ref{fig:teaser}), demonstrating its efficiency and effectiveness. We summarize the main contributions of our work as follows:
\begin{itemize}[leftmargin=*]
  \item[$\bullet$] We propose to assemble multiple zero-cost proxies for training-free NAS, assessing network architectures from various perspectives to predict a reliable ranking of the candidate architectures without training.
  \item[$\bullet$] We design novel zero-cost proxies tailored for AZ-NAS to capture distinct network characteristics, enabling a comprehensive evaluation of networks efficiently.
  \item[$\bullet$] We present a non-linear ranking aggregation method to combine rankings predicted from the proposed proxies, selecting a network highly-ranked across the proxies.
  \item[$\bullet$] We achieve the best NAS results on various search spaces, and provide extensive analyses that verify the efficacy and efficiency of AZ-NAS.
\end{itemize} 

\section{Related work} \label{sec:related_work}
NAS automates designing novel network architectures by seeking for an optimal combination of operations~\cite{ying2019bench,dong2020nasbench201,liu2019darts} and/or the width and depth of networks~\cite{sandler2018mobilenetv2,cai2019proxylessnas,lin2021zen} under limited FLOPs/\#Params budgets. Early approaches adopt reinforcement learning~\cite{zoph2018learning,zoph2017neural} or an evolutionary algorithm~\cite{real2019regularized} that require training networks iteratively, making the NAS process computationally expensive. 

To address this problem, one-/few-shot NAS methods~\cite{bender2018understanding,liu2019darts,cai2019proxylessnas,cai2020once,zhao2021few,hu2022generalizing} use parameter-shared networks, called supernets, where each sub-path in the supernets corresponds a specific network architecture in a search space. They train a single~\cite{bender2018understanding,liu2019darts,cai2019proxylessnas,cai2020once} or few~\cite{zhao2021few,hu2022generalizing} supernets, and measure the accuracies of candidate networks by sampling corresponding sub-paths from the supernets, saving the computational costs associated with an iterative training process. Sharing parameters in supernets, however, often leads to a forgetting problem~\cite{zhang2020overcoming} or biased selections~\cite{chu2021fairnas}, requiring specialized training algorithms for supernets. Moreover, constructing supernets and storing the parameters cause significant memory requirements~\cite{bender2018understanding,cai2019proxylessnas}.


Training-free NAS removes the network training process in the search phase. They use zero-cost proxies that can reflect the final performance, typically leveraging activations or gradients. Motivated by the finding that each ReLU function in a network divides an input space into two distinct pieces, called linear regions~\cite{hanin2019deep,hanin2019complexity}, the works of~\cite{chen2020tenas,mellor2021neural} examine activations to figure out the number of linear regions over the input space, which is useful for assessing the network's expressivity. These methods are however only applicable to networks employing ReLU non-linearities. ZenNAS~\cite{lin2021zen} evaluates the expressivity of networks based on an expected Gaussian complexity~\cite{kakade2008complexity}, requiring additional forward passes with perturbed inputs. Inspired by network pruning~\cite{han2015learning}, the work of~\cite{abdelfattah2021zerocost} proposes to use the pruning-at-initialization metrics~\cite{lee2019snip,wang2020picking,tanaka2020pruning} for training-free NAS, estimating the importance of each weight parameter by analyzing its gradient. TE-NAS~\cite{chen2020tenas} exploits a neural tangent kernel~(NTK)~\cite{jacot2018neural} to investigate training dynamics, and GradSign~\cite{zhang2022gradsign} attempts to search networks whose local optima obtained with different input samples are close to each other. They are relatively slower than other methods, since computing NTK~\cite{chen2020tenas} and sample-wise analysis~\cite{zhang2022gradsign} are computationally expensive, respectively. The works of~\cite{li2023zico,sun2023unleashing} analyze networks in terms of convergence and generalization by using gradient statistics computed from multiple forward and backward passes, favoring networks with gradients being large magnitudes and low variances. Despite the theoretical and empirical findings of all the aforementioned methods, the rankings of networks predicted by these methods often show weak correlations with the ground truth. For example, it has proven that most training-free NAS methods~\cite{abdelfattah2021zerocost,wang2020picking,lee2019snip,tanaka2020pruning,mellor2021neural,lin2021zen} are not effective, compared to basic proxies using FLOPs and \#Params~\cite{ning2021evaluating,li2023zico}. We conjecture that the reason for this limitation might be assessing networks with a single zero-cost proxy. We thus consider various proxies designed under distinct points of view, scoring networks comprehensively by leveraging activations, gradients, and FLOPs.

Similar to ours, the works of~\cite{chen2020tenas,abdelfattah2021zerocost,ning2021evaluating} combine zero-cost proxies to evaluate network architectures. TE-NAS~\cite{chen2020tenas} uses both the number of linear regions~\cite{hanin2019complexity,xiong2020number} and the condition number of NTK~\cite{jacot2018neural,lee2019wide} to assess expressivity and trainability of networks, respectively. However, counting linear regions is infeasible for large networks~\cite{lin2021zen}, and calculating NTK is computationally demanding~\cite{novak2022fast}, making it difficult to apply TE-NAS to search spaces for large network architectures~\cite{sandler2018mobilenetv2,cai2019proxylessnas,lin2021zen}. On the contrary, AZ-NAS remains computationally efficient while leveraging four zero-cost proxies jointly, retaining the advantage of training-free NAS. The works of~\cite{abdelfattah2021zerocost,ning2021evaluating} propose to combine a few zero-cost proxies. However, simply combining the proxies is not sufficient for leveraging complementary features among the proxies~\cite{ning2021evaluating}. For example, the work of~\cite{abdelfattah2021zerocost} chooses multiple pruning-at-initialization metrics~\cite{lee2019snip,tanaka2020pruning} for the ensemble, but they capture similar network characteristics useful for pruning, \ie, the saliency of weight parameters. Differently, we design zero-cost proxies to analyze networks from various perspectives, making them mutually complementary. We also integrate the proxies using a non-linear ranking aggregation method, assembling them effectively and boosting the NAS performance significantly.

\section{Method} \label{sec:method}
Our zero-cost proxies analyze activations and their gradients on \textit{primary blocks} of a network, where each block consists of a set of layers with different types of operations and multiple paths. For example, we consider a cell structure of NAS-Bench-201~\cite{dong2020nasbench201}, a residual block~\cite{he2016deep}, or an attention block of transformers~\cite{vaswani2017attention,dosovitskiy2020image} as a primary block. In the following, we describe the zero-cost proxies of AZ-NAS~(Sec.~\ref{sec:proxies}) and the non-linear ranking aggregation method~(Sec.~\ref{sec:aggregation}) in detail.

\begin{figure}[t]
  \captionsetup[subfigure]{justification=centering}
  \begin{center}
    \begin{subfigure}[t]{0.495\columnwidth}
      \centering
      \includegraphics[width=0.95\columnwidth]{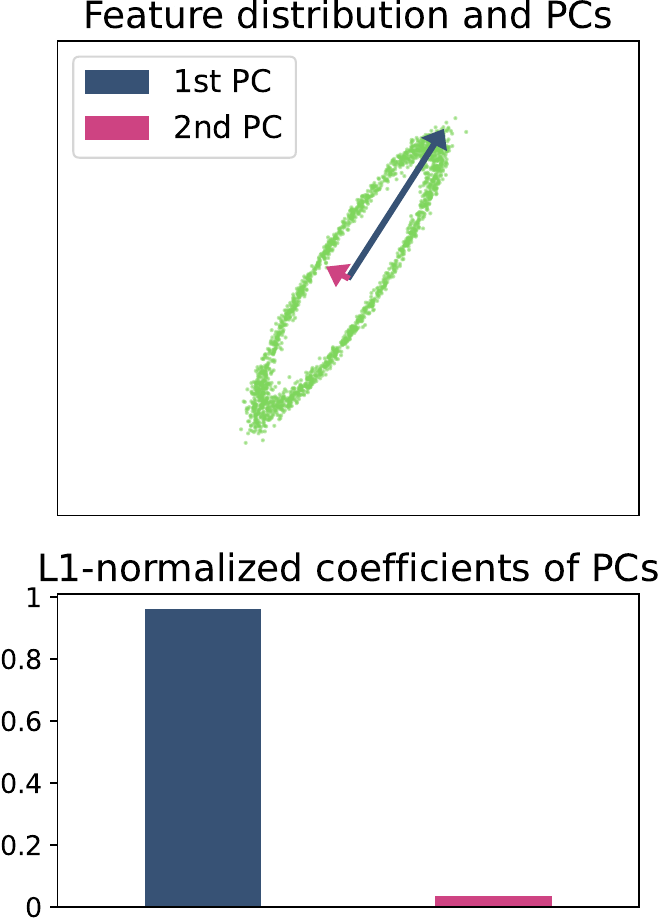}
      \caption{An example with~$s^{\mathcal{E}}=0.23$.}
      \label{fig:expressivity_2nd}
    \end{subfigure}
    \begin{subfigure}[t]{0.495\columnwidth}
      \centering
      \includegraphics[width=0.95\columnwidth]{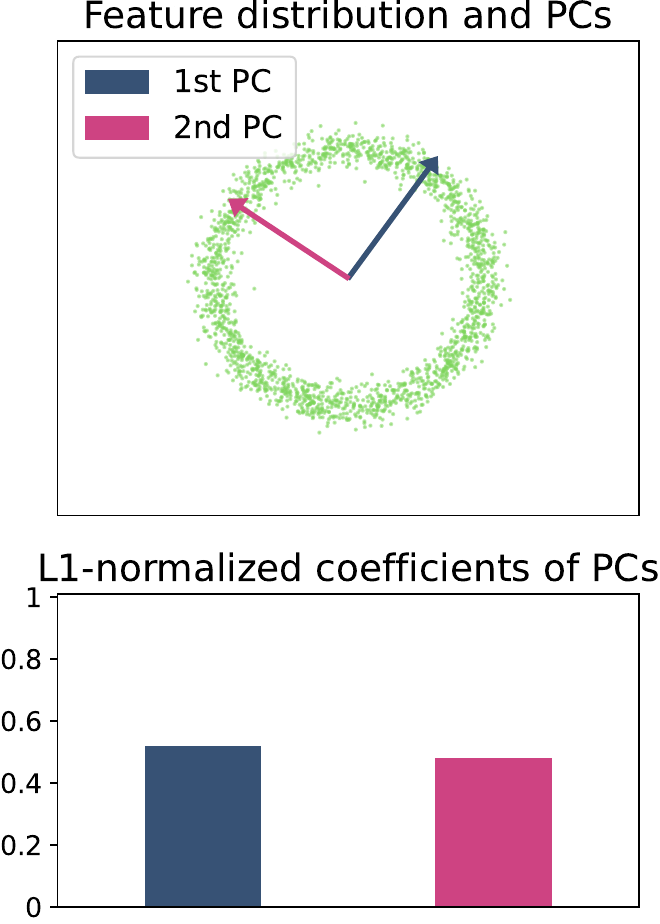}
      \caption{An example with~$s^{\mathcal{E}}=1.00$.}
      \label{fig:expressivity_1st}
    \end{subfigure}
\end{center}
        \vspace{-0.58cm}
        \caption{Toy examples for the expressivity score~$s^{\mathcal{E}}$. In (a) and (b), we synthesize 2-dimensional features~(green dots) using different covariances, and compare the L1-normalized coefficients of PCs. The features in (b) exhibit a higher expressivity score, forming an isotropic feature space.}
  \label{fig:expressivity}
  \vspace{-0.35cm}
\end{figure}

\subsection{Zero-cost proxies of AZ-NAS} \label{sec:proxies}
\paragraph{Expressivity.}
We evaluate the expressivity of a network by examining how uniformly features are distributed across the orientations in a feature space, given randomly initialized network weights and Gaussian random inputs. We refer to this as an isotropy of a feature space~\cite{cai2021isotropy}. The stronger isotropy of the feature space at initialization implies that the features are less correlated to each other~\cite{cogswell2016reducing,huang2018decorrelated,hua2021feature}. The network thus has more capacity to store various semantics during training, and it is likely to show high performance without suffering from \eg, dead neurons~\cite{lu2019dying} or a dimensional collapse problem~\cite{hua2021feature}. Motivated by principal component analysis~(PCA), we quantify the isotropy of a feature space based on the coefficients of principal components~(PCs) for the space. Specifically, we denote by~$f_{l} \in \mathbb{R}^{c \times n}$ $c$-dimensional output features of the $l$-th primary block, where $n$ is the number of features. To obtain the coefficients of PCs, we first center the features as follows:
\begin{equation}
  \vspace{-0.05cm}
  \bar{f_l}(p) = f_l (p) - \frac{1}{n} \sum_{q=1}^{n} f_l(q),
\end{equation}
where we denote by $f_l(p) \in \mathbb{R}^{c \times 1}$ the $p$-th feature vector of~$f_l$. We then compute a covariance matrix of the centered features as follows:
\begin{equation}
  \vspace{-0.1cm}
  V_l = \frac{1}{n-1} \bar{f_l}  \bar{f_l}^\top.
\end{equation}
With the covariance matrix~$V_l \in \mathbb{R}^{c \times c}$ at hand, we apply eigenvalue decomposition to obtain a set of coefficients for PCs, denoted by~$\lambda_l \in \mathbb{R}^{c \times 1}$. The coefficients are proportional to the variances of corresponding PCs, implying the importance of each PC. If a feature space is dominated by few PCs,~\eg, due to the dimensional collapse~\cite{hua2021feature}, only corresponding coefficients are large, while the others become small~(Fig.~\ref{fig:expressivity}(a)). On the other hand, if features are distributed isotropically in a space, all PCs become equally important with similar coefficients~(Fig.~\ref{fig:expressivity}(b)). Based on this, we define the expressivity score of the $l$-th primary block, $s^{\mathcal{E}}_l$, as an entropy score, posing L1-normalized coefficients of PCs as probabilities:
\begin{equation} \label{eq:partial_expressivity}
  \vspace{-0.15cm}
  s^{\mathcal{E}}_l = \sum_{i=1}^{c} - \tilde{\lambda}_l (i) \log \tilde{\lambda}_l (i),
  \vspace{-0.1cm}
\end{equation}
where we denote by~$\tilde{\lambda}_l$ a set of L1-normalized coefficients. We then obtain the expressivity score of a network $s^{\mathcal{E}}$ by summing up the block-level scores~$s^{\mathcal{E}}_l$ over the primary blocks:
\begin{equation}
  s^{\mathcal{E}} = \sum_{l=1}^{L} s^{\mathcal{E}}_l,
\end{equation}
where $L$ is the total number of primary blocks in a network. The expressivity proxy is particularly useful for detecting redundancy in a network, such as reducible channel dimensions caused by dead neurons.

\vspace{-0.35cm}
\paragraph{Progressivity.}
The widths of modern network architectures~\cite{he2016deep,sandler2018mobilenetv2,tan2019efficientnet} gradually increase for deeper blocks, making it possible to capture high-level semantics through deep features with large capacities. Building upon this, we propose a progressivity proxy that evaluates an ability to expand a feature space progressively according to the depth of primary blocks. We define the progressivity score~$s^{\mathcal{P}}$ using the difference of the block-level expressivity scores in~Eq.~\eqref{eq:partial_expressivity}:
\begin{equation}
  s^{\mathcal{P}} = \min_{l \in \{ 2, \cdots, L \}} s^{\mathcal{E}}_{l} - s^{\mathcal{E}}_{l-1},
\end{equation}
which computes the smallest difference in expressivity scores between neighboring primary blocks. A high progressivity score of a network indicates that its block-level expressivity scores consistently increase, at least by the value of the progressivity score, along the depth of the primary blocks. Our progressivity proxy automates the NAS process by considering how features evolve throughout a forward pass explicitly in terms of expressivity, without requiring strict and manual constraints on the width of a network~\eg, as in~\cite{shen2023deepmad}.

\vspace{-0.35cm}
\paragraph{Trainability.}
The seminal work of~\cite{miyato2018spectral} has shown that gradients can be backpropagated without diverging or vanishing, only when the spectral norm of a Jacobian matrix for each layer is close to $1$. Motivated by this, we design a trainability score with the spectral norm of a Jacobian matrix. This evaluates the stable propagation of gradients at initialization, which has proven to be crucial for high-performing networks~\cite{glorot2010understanding,he2015delving,qi2023lipsformer}. However, different from the previous work~\cite{miyato2018spectral} dealing with a simple network architecture, we should consider complicated structures of primary blocks consisting of various operations and multiple paths across layers~(\eg, a cell structure~\cite{dong2020nasbench201}). In this case, obtaining a Jacobian matrix is not a trivial task. To circumvent this issue, we devise an efficient approximation of the Jacobian matrix of a primary block, and propose to use the approximation to compute the trainability score. We first simplify the forward pass~$\psi_l$ of the~$l$-th primary block by approximating it as a linear system that inputs $c^\prime$-dimensional features~$f_{l-1} \in \mathbb{R}^{c^\prime \times n}$ and outputs $c$-dimensional features~$f_{l} \in \mathbb{R}^{c \times n}$:
\begin{equation}
  f_{l} = \psi_l (f_{l-1}) \approx A_{l} f_{l-1},
\end{equation}
where~$A_l \in \mathbb{R}^{c \times c^\prime}$ is a linear transformation matrix. Under this approximation, the backward pass~$\phi_l$ of the primary block can be represented as follows:
\begin{equation} \label{eq:backpropagation} 
  g_{l-1} = \phi_l (g_{l}) \approx A_{l}^{\top} g_{l},
\end{equation} 
where~$g_{l-1} \in \mathbb{R}^{c^\prime \times n}$ and~$g_{l} \in \mathbb{R}^{c \times n}$ are the gradients of~$f_{l-1}$ and~$f_{l}$, respectively, and the transformation matrix~$A_{l}$ serves as a Jacobian matrix. Motivated by the Hutchinson's method~\cite{avron2011randomized} used in~\cite{yao2020pyhessian,dong2020hawq,lee2021network}, we introduce a Rademacher random vector~$v \in \mathbb{R}^{c \times 1}$, whose elements are randomly drawn from $\{-1, 1\}$ with a equal probability, to obtain the Jacobian matrix~$A_{l}$ in the approximation. Specifically, we rewrite Eq.~\eqref{eq:backpropagation} by substituting the output gradient with the Rademacher random vector~$v$:
\begin{equation} \label{eq:backpropagation2}
  u = \phi_l (v) \approx A_{l}^{\top} v,
\end{equation}
where~$u \in \mathbb{R}^{c^\prime \times 1}$ is considered as a gradient propagated from~$v$. By using the property that~$\mathbb{E} \left[ v v^{\top} \right]$ is an identity matrix~\cite{yao2020pyhessian,lee2021network}, we have the following relationship:
\begin{equation}
  \mathbb{E} \left[ u v^{\top} \right] \approx \mathbb{E} \left[A_{l}^{\top} v v^{\top} \right] = A_{l}^{\top} \mathbb{E} \left[ v v^{\top} \right] = A_{l}^{\top}.
\end{equation}
That is, if we can compute~$u$ in Eq.~\eqref{eq:backpropagation2} by feeding the Rademacher random vector~$v$ into the function~$\phi_l$, we can compute~$A_{l}^{\top}$ using~$\mathbb{E} \left[ u v^{\top} \right]$. Based on this, we compute the Jacobian matrix~$A_{l}$ in the approximation of~Eq.~\eqref{eq:backpropagation} as follows:
\begin{equation}
  A_{l}^{\top} = \frac{1}{n} \sum_{p=1}^{n} g_{l-1}(p) g_{l}^{\top}(p),
\end{equation}
where we synthesize the output gradients~$g_{l}$ with Rademacher random vectors and obtain the input gradients~$g_{l-1}$ by an automatic backpropagation tool~\cite{paszke2019pytorch}. Consequently, we define the trainability score~$s^{\mathcal{T}}$ as follows:
\begin{equation} \label{eq:trainability}
  s^{\mathcal{T}} = \frac{1}{L-1} \sum_{l=2}^{L} -\sigma_{l} - \frac{1}{\sigma_{l}} + 2,
\end{equation}
which has a maximum value, when the spectral norm of~$A_{l}$, denoted by $\sigma_{l}$, is equal to $1$ for all~$l$. In Eq.~\eqref{eq:trainability}, both the spectral norm~$\sigma_{l}$ and its reciprocal value contribute equally to the score. This penalizes, for example, the spectral norms~$\sigma_{l}$ of~$2$ and $\frac{1}{2}$ to the same extent, equivalently treating the increasing and decreasing scale differences relative to $1$. The proposed trainability proxy provides distinct advantages compared to previous zero-cost proxies using gradients. First, it is relatively faster than others~\cite{wang2020picking,lee2019snip,tanaka2020pruning,chen2020tenas,zhang2022gradsign,li2023zico}, and the score can be obtained simultaneously with other proxy scores of AZ-NAS~(\ie, expressivity and progressivity) with the same input sample. Second, our trainability proxy is not tied to specific operations, since it computes the score using input and output gradients of any types of primary blocks. In contrast, existing gradient-based proxies~\cite{abdelfattah2021zerocost,wang2020picking,lee2019snip,tanaka2020pruning,chen2020tenas,zhang2022gradsign,li2023zico} exploit gradients of trainable weight parameters, suggesting that they cannot evaluate the blocks consisting of non-parametric operations~(\eg, a cell with average pooling) directly.

\vspace{-0.35cm}
\paragraph{Complexity.}
Recent studies~\cite{ning2021evaluating,li2023zico} have shown that architectural characteristics related to hardware resources, such as \#Params or FLOPs, are correlated closely with the performance of networks. Following this observation, we propose to use FLOPs itself as a complexity score~$s^{\mathcal{C}}$, preferring the networks that maximally use computational resources within a given budget.

\subsection{Non-linear ranking aggregation} \label{sec:aggregation}
A straightforward way for integrating the proxies into the final scores would be a linear summation of rankings~\cite{chen2020tenas}. If we have both lowly- and highly-ranked proxies from a network, the linear aggregation makes it difficult to highlight an undesirable characteristic captured by the lowly-ranked proxy. This is because the low ranking of the proxy could be easily offset by the highly-ranked proxies. We use a non-linear ranking aggregation method to combine our proxies more effectively. Specifically, we denote by~$\mathbb{S}^{\mathcal{M}}=[ s^{\mathcal{M}} (1), \cdots, s^{\mathcal{M}} (m) ]$ a set of scores for the proxy~$\mathcal{M} \in \{ \mathcal{E}, \mathcal{P}, \mathcal{T}, \mathcal{C} \}$, obtained with~$m$ candidate architectures. The non-linear ranking aggregation outputs the final AZ-NAS score for the~$i$-th network~$s^{\textrm{AZ}}(i)$ as follows:
\begin{equation}
  s^{\textrm{AZ}} (i) = \sum_{\mathcal{M} \in \{ \mathcal{E},~\mathcal{P},~\mathcal{T},~\mathcal{C} \}}  \log{\frac{\textrm{Rank} \left( s^{\mathcal{M}} (i) \right)}{m}},
\end{equation}
where $\textrm{Rank} (\cdot)$ assigns the ranking of an input score~$s^{\mathcal{M}} (i)$ over the scores in the set~$\mathbb{S}^{\mathcal{M}}$ in ascending order. It penalizes the final score more severely when one of the proxies indicates subpar performance, allowing us to find a network highly-ranked across all the proxies evenly. As will be shown in the ablation study~(Sec.~\ref{sec:discussion}), this approach substantially boosts the ranking consistency between the final AZ-NAS scores and the performance, compared to the linear aggregation of rankings. When we perform an evolutionary search, we apply the non-linear ranking aggregation method to the proxy scores obtained with all candidate architectures. For generating a new candidate during the search, we mutate one of the top-$k$ network architectures in terms of the final AZ-NAS scores~(Algorithm~\ref{alg:algorithm}).

\begin{algorithm}[t]
  \caption{Evolutionary search using AZ-NAS.} \label{alg:algorithm}
  \textbf{Input}: search space~$Z$; number of NAS iteration~$T$; computational budget~$B$; number of top-scoring architectures for mutation~$k$. \\
  \textbf{Output}: selected architecture~$F^\ast$.
  
  \begin{algorithmic}[1]
     \STATE Initialize the first architecture~$F_1$ for an evolutionary search.
     \STATE Initialize empty history sets for storing architectures~$\mathbb{F}$ and proxy scores~$\mathbb{S^{\mathcal{E}}}$,~$\mathbb{S^{\mathcal{P}}}$,~$\mathbb{S^{\mathcal{T}}}$, and~$\mathbb{S^{\mathcal{C}}}$.
     \STATE {\textbf{while} $i = 1$ \textbf{to} $T$ \textbf{do}}
     \STATE \hspace{\algorithmicindent}Compute the proxy scores~$s^{\mathcal{E}}$,~$s^{\mathcal{P}}$,~$s^{\mathcal{T}}$, and~$s^{\mathcal{C}}$ of the\\\hspace{\algorithmicindent}architecture $F_i$.
     \STATE \hspace{\algorithmicindent}Append the architecture~$F_i$ and the scores~$s^{\mathcal{E}}$, $s^{\mathcal{P}}$,\\\hspace{\algorithmicindent}$s^{\mathcal{T}}$, $s^{\mathcal{C}}$ to the corresponding history sets~$\mathbb{F}$, $\mathbb{S^{\mathcal{E}}}$, $\mathbb{S^{\mathcal{P}}}$,\\\hspace{\algorithmicindent}$\mathbb{S^{\mathcal{T}}}$, and $\mathbb{S^{\mathcal{C}}}$, respectively.
     \STATE \hspace{\algorithmicindent}Compute the AZ-NAS scores of the architectures in\\\hspace{\algorithmicindent}$\mathbb{F}$ using the non-linear ranking aggregation with the\\\hspace{\algorithmicindent}proxy scores in~$\mathbb{S^{\mathcal{E}}}$,~$\mathbb{S^{\mathcal{P}}}$,~$\mathbb{S^{\mathcal{T}}}$, and~$\mathbb{S^{\mathcal{C}}}$.
     \STATE \hspace{\algorithmicindent}Generate a new candidate architecture~$F_{i+1}$ belong-\\\hspace{\algorithmicindent}ing to the search space~$Z$ by mutating one of the top-\\\hspace{\algorithmicindent}$k$ architectures based on the AZ-NAS scores, within\\\hspace{\algorithmicindent}the computational budget~$B$.
     \STATE {\textbf{end while}}
     \STATE Select the architecture~$F^\ast$ showing the highest AZ-NAS score among the ones in the history set~$\mathbb{F}$.
  \end{algorithmic}
\end{algorithm}

\section{Experiment} \label{sec:experiment}
In this section, we describe experimental settings~(Sec.~\ref{sec:exp_details}) and compare AZ-NAS with the state of the art~(Sec.~\ref{sec:quantitative}). We then present an ablation study and an analysis on integrating other zero-cost proxies~(Sec.~\ref{sec:discussion}). More results can be found in the supplement.

\begin{table*}[t]
  \small
  \setlength{\tabcolsep}{0.25cm}
  \centering
  \caption{Quantitative comparison of the training-free NAS methods on NAS-Bench-201~\cite{dong2020nasbench201}. We categorize the types of zero-cost proxies into architectural (A), backward (B), and forward (F) ones depending on the inputs of the proxies. We report Kendall's $\tau$~(KT) and Spearman's $\rho$~(SPR) computed with all candidate architectures, together with an average runtime. We also provide the average and standard deviation of test accuracies~(Acc.) on each dataset, where they are obtained through 5 random runs. To this end, we randomly sample 3000 candidate architectures for each run and share the same architecture sets across all the methods. All results are reproduced with the official codes provided by the authors.} 
  \vspace{-0.3cm}
  \begin{adjustbox}{max width=\textwidth}
     \begin{tabular}{l c c@{\hspace{0.25cm}}c@{\hspace{0.25cm}}c c@{\hspace{0.25cm}}c@{\hspace{0.25cm}}c c@{\hspace{0.25cm}}c@{\hspace{0.25cm}}c S[table-format=4.1, detect-weight]}
        \toprule
        \multirow{2}{*}[-2pt]{Method} & \multirow{2}{*}[-2pt]{Type} & \multicolumn{3}{c}{CIFAR-10}     & \multicolumn{3}{c}{CIFAR-100}    & \multicolumn{3}{c}{ImageNet16-120}  & {\multirow{2}{*}[-2pt]{\shortstack[c]{Runtime \\  (ms/arch) }}} \\ \cmidrule(lr){3-5} \cmidrule(lr){6-8} \cmidrule(lr){9-11}
                                                                 &   & KT & SPR & Acc.                                      & KT & SPR & Acc.                                & KT & SPR & Acc.                      &  \\
        \midrule 
        \#Params                                                 & A     & 0.578     & 0.753     & 93.50 $\pm$ \U{0.17}     & 0.552     & 0.728     & \U{70.63} $\pm$ 0.29   & 0.520     & 0.691     & 41.91 $\pm$ \U{0.70}     & {-}  \\
        FLOPs                                                    & A     & 0.578     & 0.753     & 93.50 $\pm$ \U{0.17}     & 0.551     & 0.727     & \U{70.63} $\pm$ 0.29   & 0.517     & 0.691     & 41.91 $\pm$ \U{0.70}     & {-}  \\
        GradNorm~\cite{abdelfattah2021zerocost}                  & B     & 0.357     & 0.484     & 90.13 $\pm$ 1.00         & 0.350     & 0.475     & 62.95 $\pm$ 1.38       & 0.304     & 0.412     & 24.77 $\pm$ 4.67         & 28.8 \\ 
        Grasp~\cite{abdelfattah2021zerocost,wang2020picking}     & B     & 0.318     & 0.460     & 89.49 $\pm$ 1.63         & 0.315     & 0.453     & 61.72 $\pm$ 2.97       & 0.282     & 0.406     & 27.26 $\pm$ 4.92         & 395.9  \\
        Snip~\cite{abdelfattah2021zerocost,lee2019snip}          & B     & 0.454     & 0.615     & 90.32 $\pm$ 1.39         & 0.451     & 0.609     & 63.44 $\pm$ 2.52       & 0.400     & 0.539     & 28.77 $\pm$ 5.74         & 326.5  \\
        Synflow~\cite{abdelfattah2021zerocost,tanaka2020pruning} & B     & 0.571     & 0.769     & 93.06 $\pm$ 0.83         & 0.565     & 0.761     & 69.24 $\pm$ 2.07       & 0.555     & 0.747     & 40.51 $\pm$ 5.70         & 53.4  \\
        NASWOT~\cite{mellor2021neural}                           & F     & 0.557     & 0.743     & 92.67 $\pm$ 0.24         & 0.579     & 0.769     & 69.91 $\pm$ \U{0.27}   & 0.573     & 0.760     & \U{44.45} $\pm$ 0.83     & 36.9  \\
        TE-NAS~\cite{chen2020tenas}                              & B+F   & 0.536     & 0.731     & 92.30 $\pm$ 0.33         & 0.535     & 0.728     & 67.87 $\pm$ 1.05       & 0.492     & 0.680     & 40.49 $\pm$ 2.16         & 1311.8  \\
        ZenNAS~\cite{lin2021zen}                                 & F     & 0.296     & 0.385     & 90.30 $\pm$ 0.36         & 0.283     & 0.361     & 67.70 $\pm$ 1.00       & 0.303     & 0.409     & 39.55 $\pm$ 1.27         & 19.9 \\ 
        GradSign~\cite{zhang2022gradsign}                        & B     & \U{0.618} & \U{0.809} & \U{93.52} $\pm$ 0.19     & \U{0.594} & \U{0.784} & 70.57 $\pm$ 0.31       & 0.575     & 0.765     & 41.89 $\pm$ 0.69         & 1823.9  \\
        ZiCo~\cite{li2023zico}                                   & B     & 0.589     & 0.784     & 93.50 $\pm$ 0.18         & 0.590     & 0.785     & 70.62 $\pm$ \B{0.26}   & \U{0.584} & \U{0.778} & 42.04 $\pm$ 0.82         & 372.8  \\
        AZ-NAS~(Ours)                                            & A+B+F & \B{0.741} & \B{0.913} & \B{93.53} $\pm$ \B{0.15} & \B{0.723} & \B{0.900} & {\B{70.75}} $\pm$ 0.48 & \B{0.710} & \B{0.886} & \B{45.43} $\pm$ \B{0.29} & 42.7  \\
        \midrule
        Ground truth                                             & -     & -         & -         & 94.29 $\pm$ 0.13         & -         & -         & 73.25 $\pm$ 0.26       & -         & -         & 47.05 $\pm$ 0.30         & {-} \\
        \bottomrule
     \end{tabular} \label{tab:NB201}
  \end{adjustbox}
  \vspace{-0.4cm}
\end{table*}

\subsection{Experimental settings} \label{sec:exp_details}
\vspace{-0.1cm}
We perform extensive experiments on the NAS-Bench-201 \cite{dong2020nasbench201}, MobileNetV2~\cite{sandler2018mobilenetv2,lin2021zen} and AutoFormer~\cite{chen2021autoformer} search spaces. During the search phase, we measure the proxy scores of AZ-NAS using a single batch of Gaussian random inputs with a batch size of 64. In the following, we describe experimental settings for each search space in detail.

\vspace{-0.35cm}
\paragraph{NAS-Bench-201.}
NAS-Bench-201~\cite{dong2020nasbench201} consists of 15625 network architectures, each of which uses a unique cell structure. It provides the diagnostic information~\eg, test accuracies and training losses on CIFAR-10/100~\cite{krizhevsky2009learning} and ImageNet16-120~(IN16-120)~\cite{chrabaszcz2017downsampled}. We measure the correlation coefficients in terms of Kendall's~$\tau$ and Spearman's~$\rho$ between predicted and ground-truth network rankings across all the candidate architectures. We also report the top-1 test accuracies of selected networks, averaged over 5 runs, together with standard deviations. For each run, we randomly sample 3000 architectures, and discover the best one based on the AZ-NAS score.

\vspace{-0.35cm}
\paragraph{MobileNetV2.}
We follow the configuration of the MobileNetV2~\cite{sandler2018mobilenetv2} search space proposed in~\cite{lin2021zen} for a fair comparison with previous works~\cite{lin2021zen,li2023zico}. This space includes candidate architectures built with inverted residual blocks~\cite{sandler2018mobilenetv2}, with varying depth, width, and expansion ratio of the block. We perform an evolutionary search in Algorithm~\ref{alg:algorithm} to find the optimal networks, while setting the number of iterations~$T$ and~$k$ to 1e5 and 1024, respectively. Following~\cite{li2023zico}, we search for networks under FLOPs constraints of 450M, 600M, and 1000M. We train the selected networks on ImageNet~\cite{deng2009imagenet} using the same training setting as in~\cite{lin2021zen,li2023zico}, and report the top-1 validation accuracy.

\vspace{-0.35cm}
\paragraph{AutoFormer.}
The AutoFormer~\cite{chen2021autoformer} search space is designed to evaluate the NAS methods specialized to Vision Transformers~(ViTs)\footnote{We do not apply the progressivity proxy for the AutoFormer search space, since we empirically find that attention modules in ViTs produce similar attention values across tokens with Gaussian random inputs at initialization. In this case, output features of the attention modules thus tend to become close in a feature space, which could not guarantee the feature space to be expanded along the depth. Developing zero-cost proxies suitable for ViTs could enhance the NAS performance of AZ-NAS for ViTs, and we leave this for a future work.}.~It contains ViT architectures with variable factors of depth, embedding dimension, number of attention heads, and expansion ratio for multi-layer perceptrons. The search space is further divided into the Tiny, Small, and Base subsets according to the model sizes. We select a ViT architecture with the best AZ-NAS score for each subset, among 10000 randomly chosen architectures, similar to~\cite{zhou2022training}. We train and evaluate the selected ViTs on ImageNet following the configuration in~\cite{chen2021autoformer}, except for the one from the Base subset, for which we reduce the number of training epochs as in~\cite{zhou2022training} to avoid overfitting.

\subsection{Results} \label{sec:quantitative}
\vspace{-0.1cm}
\paragraph{NAS-Bench-201.}
We show in Table~\ref{tab:NB201} a quantitative comparison between AZ-NAS and the state-of-the-art training-free NAS methods on NAS-Bench-201~\cite{dong2020nasbench201}. We report correlation coefficients between predicted and ground-truth rankings of networks, and top-1 test accuracies of selected networks. We can see that AZ-NAS achieves the best ranking consistency w.r.t the performance in terms of Kendall's~$\tau$ and Spearman's~$\rho$, outperforming others by significant margins across all the datasets. This enables discovering networks that consistently show better performance compared to the networks chosen from other methods. The training-free NAS methods in Table~\ref{tab:NB201} focus on capturing a single network characteristic using either activations or gradients only, except for TE-NAS~\cite{chen2020tenas} requiring lots of computational costs. On the contrary, AZ-NAS examines networks from various perspectives based on activations, gradients, and FLOPs, providing a comprehensive evaluation of networks. Moreover, the proposed proxies tailored for AZ-NAS are computationally efficient and can be computed simultaneously within a single forward and backward pass. Note that several training-free NAS methods exploit specialized techniques for implementation, such as removing batch normalization~\cite{tanaka2020pruning}, analyzing activations associated only with ReLU non-linearities~\cite{chen2020tenas,mellor2021neural}, or iterating multiple forward and/or backward passes~\cite{chen2020tenas,lin2021zen,zhang2022gradsign,li2023zico}. In contrast, AZ-NAS performs without additional complex operations or modifications to network architectures, assessing the networks efficiently without bells and whistles.

\begin{table}[t]
  \renewcommand*{\arraystretch}{0.95}
  \small
  \setlength{\tabcolsep}{0.35em}
  \centering
  \caption{Quantitative comparison of networks chosen by the NAS methods on ImageNet~\cite{deng2009imagenet} in terms of the top-1 validation accuracy. We group the networks with similar FLOPs~(\ie, 450M, 600M, and 1000M) and categorize the NAS methods into multi-shot~(MS), one-shot~(OS), and zero-shot~(ZS) ones, according to the number of networks to train during the search phase. For each FLOPs constraint, we report the results of AZ-NAS averaged over three random runs starting from the search phase. All numbers except ours are taken from~\cite{li2023zico}.}
  \vspace{-0.3cm}
  \begin{adjustbox}{max width=\columnwidth}
     \begin{tabular}{l c c c c}
        \toprule
        \multirow{2}{*}{Method} & \multirow{2}{*}{FLOPs} & \multirow{2}{*}{Top-1 acc.} & \multirow{2}{*}{Type} & \multirow{2}{*}{\shortstack[c]{Search cost \\ (GPU days)}} \\
        \\
        \midrule \addlinespace[-0.001cm]
        \multicolumn{5}{c}{450M} \\ \addlinespace[-0.08cm]
        \midrule
        NASNet-B~\cite{zoph2018learning} & 488M            & 72.8                 & MS & 1800 \\
        CARS-D~\cite{yang2020cars}       & 496M            & 73.3                 & MS & 0.4 \\
        BN-NAS~\cite{chen2021bn}         & 470M            & 75.7                 & MS & 0.8 \\
        OFA~\cite{cai2020once}           & 406M            & 77.7                 & OS & 50 \\
        RLNAS~\cite{zhang2021neural}     & 473M            & 75.6                 & OS & - \\
        DONNA~\cite{moons2021distilling} & 501M            & 78.0                 & OS & 405 \\
        \# Params                        & 451M            & 63.5                 & ZS & 0.02 \\
        ZiCo~\cite{li2023zico}           & 448M            & 78.1                 & ZS & 0.4 \\
        AZ-NAS~(Ours)                    & 462M $\pm$ 1.5M & {\B{78.6}} $\pm$ 0.2 & ZS & 0.4 \\
        \midrule \addlinespace[-0.001cm]
        \multicolumn{5}{c}{600M} \\ \addlinespace[-0.08cm]
        \midrule
        PNAS~\cite{liu2018progressive}          & 588M            & 74.2                 & MS & 224 \\ 
        CARS-I~\cite{yang2020cars}              & 591M            & 75.2                 & MS & 0.4 \\ 
        DARTS~\cite{liu2019darts}               & 574M            & 73.3                 & OS & 4 \\ 
        ProxylessNAS~\cite{cai2019proxylessnas} & 595M          & 76.0                 & OS & 8.3 \\ 
        OFA~\cite{cai2020once}                  & 662M            & 78.7                 & OS & 50 \\ 
        RLNAS~\cite{zhang2021neural}            & 597M            & 75.9                 & OS & - \\ 
        DONNA~\cite{moons2021distilling}        & 599M            & 78.4                 & OS & 405 \\ 
        ZenNAS~\cite{lin2021zen}                & 611M            & 79.1                 & ZS & 0.5 \\
        ZiCo~\cite{li2023zico}                  & 603M            & 79.4                 & ZS & 0.4 \\
        AZ-NAS~(Ours)                           & 615M $\pm$ 2.2M & {\B{79.9}} $\pm$ 0.3 & ZS & 0.6 \\
        \midrule \addlinespace[-0.001cm]
        \multicolumn{5}{c}{1000M} \\ \addlinespace[-0.08cm]
        \midrule
        sharpDARTS~\cite{hundt2019sharpdarts} & 950M             & 76.0           & OS & - \\
        ZenNAS~\cite{lin2021zen}              & 934M             & 80.8           & ZS & 0.5 \\
        ZiCo~\cite{li2023zico}                & 1005M            & 80.5           & ZS & 0.4 \\
        AZ-NAS~(Ours)                         & 1022M $\pm$ 5.1M & {\B{81.1}} $\pm$ 0.1 & ZS & 0.7 \\
        \bottomrule
     \end{tabular} \label{tab:MBV2}
  \end{adjustbox}
  \vspace{-0.5cm}
\end{table}

\vspace{-0.4cm}
\paragraph{MobileNetV2.}
We compare in Table~\ref{tab:MBV2} the performance of various NAS methods in terms of the top-1 validation accuracy on ImageNet~\cite{deng2009imagenet}. We can observe from the table that AZ-NAS provides the best NAS results, achieving the highest accuracies across all FLOPs constraints. It even outperforms the training-based NAS approaches~(MS and OS) typically requiring expensive search costs, demonstrating its effectiveness and efficiency. AZ-NAS also provides high-performing networks consistently over multiple random runs with different seed numbers. Note that the previous training-free NAS methods~\cite{lin2021zen,li2023zico} apply an evolutionary search using MobileNetV2-styled networks, while removing residual connections. They focus on analyzing vanilla convolutional neural networks~\cite{lin2021zen} during the search, where multiple convolutional layers are simply stacked without additional connections. The residual connections are then restored to train selected networks, potentially leading to a discrepancy between the expected and final performance at the search and training phases, respectively. In contrast, AZ-NAS does not rely on such techniques, improving the practicality and avoiding the discrepancy issue. We additionally provide in the supplement a comprehensive comparison with the training-free NAS methods~\cite{lin2021zen,li2023zico} in terms of reproducibility and a search cost, under the fair settings for network search and training.


\begin{table}[t]
  \small
  \setlength{\tabcolsep}{0.35em}
  \centering
  \caption{Quantitative comparison for the AutoFormer~\cite{chen2021autoformer} search space. We report top-1 validation accuracies of selected networks on ImageNet~\cite{deng2009imagenet}. All numbers are taken from corresponding papers, except for \#Params and FLOPs of TF-TAS~\cite{zhou2022training}. We measure these numbers using the same source code as AutoFormer for a fair comparison.}
  \vspace{-0.3cm}
  \begin{adjustbox}{max width=\columnwidth}
     \begin{tabular}{l c c c c c}
        \toprule
        \multirow{2}{*}{Method} & \multirow{2}{*}{\#Params} & \multirow{2}{*}{FLOPs} & \multirow{2}{*}{Top-1 acc.} & \multirow{2}{*}{Type} & \multirow{2}{*}{\shortstack[c]{Search cost \\ (GPU days)}} \\
        \\
        \midrule \addlinespace[-0.001cm]
        \multicolumn{6}{c}{Tiny} \\ \addlinespace[-0.08cm]
        \midrule
        AutoFormer~\cite{chen2021autoformer}   & 5.70M & 1.30G & 74.7     & OS & 24 \\
        AZ-NAS~(Ours)                          & 5.92M & 1.38G & \B{76.1} & ZS & 0.03 \\ \rowcolor{Gainsboro!60}
        TF-TAS~\cite{zhou2022training}         & 6.20M & 1.43G & 75.3     & ZS & 0.5 \\ \rowcolor{Gainsboro!60}
        AZ-NAS~(Ours)                          & 6.16M & 1.43G & \B{76.4} & ZS & 0.04 \\
        \midrule \addlinespace[-0.001cm]
        \multicolumn{6}{c}{Small} \\ \addlinespace[-0.08cm]
        \midrule
        AutoFormer~\cite{chen2021autoformer}   & 22.9M & 5.10G & 81.7     & OS & 24 \\
        AZ-NAS~(Ours)                          & 23.0M & 4.94G & \B{82.0} & ZS & 0.06 \\ \rowcolor{Gainsboro!60}
        TF-TAS~\cite{zhou2022training}         & 23.9M & 5.16G & 81.9     & ZS & 0.5 \\ \rowcolor{Gainsboro!60}
        AZ-NAS~(Ours)                          & 23.8M & 5.13G & \B{82.2} & ZS & 0.07 \\
        \midrule \addlinespace[-0.001cm]
        \multicolumn{6}{c}{Base} \\ \addlinespace[-0.08cm]
        \midrule
        AutoFormer~\cite{chen2021autoformer}   & 54.0M & 11.0G & \B{82.4} & OS & 24 \\
        AZ-NAS~(Ours)                          & 53.7M & 11.4G & 82.1     & ZS & 0.11 \\ \rowcolor{Gainsboro!60}
        TF-TAS~\cite{zhou2022training}         & 56.5M & 11.9G & 82.2     & ZS & 0.5 \\ \rowcolor{Gainsboro!60}
        AZ-NAS~(Ours)                          & 54.1M & 11.5G & \B{82.3} & ZS & 0.17 \\
        \bottomrule
     \end{tabular} \label{tab:AutoFormer}
  \end{adjustbox}
  \vspace{-0.35cm}
\end{table}

\vspace{-0.35cm}
\paragraph{AutoFormer.}
To validate the generalization ability of AZ-NAS on ViT architectures, we perform experiments on the AutoFormer~\cite{chen2021autoformer} search space. We provide in Table~\ref{tab:AutoFormer} a quantitative comparison to the state-of-the-art NAS methods~\cite{chen2021autoformer,zhou2022training}, specially designed for ViTs, in terms of the validation accuracy on ImageNet~\cite{deng2009imagenet}. For each subset of the search space~(\ie, Tiny, Small, and Base), we find two ViTs with the \#Params constraints similar to AutoFormer \cite{chen2021autoformer} and TF-TAS~\cite{zhou2022training}, separately, for a fair comparison. We can see that AZ-NAS discovers ViTs showing better performance compared to other methods in most cases, with significantly lower search costs. This suggests that our method is not limited to a specific type of network architecture, demonstrating the generalization ability of our proxies.

\subsection{Discussion} \label{sec:discussion}

\begin{table}[t]
  \small
  \setlength{\tabcolsep}{0.12cm}
  \centering
  \caption{Ablation study for the zero-cost proxies of AZ-NAS on NAS-Bench-201~\cite{dong2020nasbench201}. When multiple proxies are used, we integrate them into the final scores using either the linear (L) or non-linear (NL) ranking aggregation~(RA) methods. We compare the ranking consistency w.r.t the performance in terms of Kendall's~$\tau$.}
  \vspace{-0.22cm}
  \begin{adjustbox}{max width=\columnwidth}
     \begin{tabular}{c | C{0.55cm} C{0.55cm} C{0.55cm} C{0.55cm} | C{0.75cm} | c c c}
        \toprule
        & $s^{\mathcal{E}}$ & $s^{\mathcal{P}}$ & $s^{\mathcal{T}}$ & $s^{\mathcal{C}}$ & RA & CIFAR-10 & CIFAR-100 & IN16-120 \\
        \midrule
        \circnum{1} & \cmark &        &        &        & -  & 0.569     & \B{0.563} & 0.506 \\
        \circnum{2} &        & \cmark &        &        & -  & 0.521     & 0.508     & 0.489 \\
        \circnum{3} &        &        & \cmark &        & -  & 0.349     & 0.353     & 0.407 \\
        \circnum{4} &        &        &        & \cmark & -  & \B{0.578} & 0.551     & \B{0.517} \\
        \midrule
        \circnum{5} & \cmark & \cmark &        &        & NL & 0.601     & 0.590     & 0.547 \\
        \circnum{6} & \cmark &        & \cmark &        & NL & 0.635     & 0.631     & 0.639 \\
        \circnum{7} & \cmark &        &        & \cmark & NL & \B{0.674} & \B{0.653} & 0.601 \\ %
        \circnum{8} &        &        & \cmark & \cmark & NL & 0.629     & 0.615     & \B{0.644} \\
        \midrule
        \circnum{9} & \cmark & \cmark & \cmark &        & NL & 0.679     & 0.673     & 0.669 \\
        \circnum{10}& \cmark & \cmark &        & \cmark & NL & 0.683     & 0.661     & 0.616 \\ %
        \circnum{11}& \cmark &        & \cmark & \cmark & NL & \B{0.731} & \B{0.714} & \B{0.708} \\
        \circnum{12}& \cmark &        & \cmark & \cmark & L  & 0.706     & 0.692     & 0.678 \\
        \midrule
        \circnum{13}& \cmark & \cmark & \cmark & \cmark & NL & \B{0.741} & \B{0.723} & \B{0.710} \\
        \circnum{14}& \cmark & \cmark & \cmark & \cmark & L  & 0.697     & 0.681     & 0.663 \\
        \bottomrule
     \end{tabular} \label{tab:ablation}
  \end{adjustbox}
  \vspace{-0.3cm}
\end{table}

\vspace{-0.1cm}
\paragraph{Ablation study.}
We provide in Table~\ref{tab:ablation} an ablation study on each component of AZ-NAS. We evaluate the NAS performance on NAS-Bench-201~\cite{dong2020nasbench201} using various combinations of the proposed proxies and the ranking aggregation methods. We summarize our findings as follows: (1)~We can see from \circnum{1} to \circnum{4} in Table~\ref{tab:ablation} that exploiting a single proxy does not provide satisfactory results. Aggregating more proxies improves the NAS performance substantially, outperforming most state-of-the-art NAS methods in Table~\ref{tab:NB201}, even when two of the proxies in AZ-NAS are selected in~\circnum{5} to \circnum{8}. These results demonstrate the effectiveness of assembling various zero-cost proxies, confirming the importance of a comprehensive evaluation from various perspectives for training-free NAS. (2)~By comparing \circnum{11} and \circnum{12} (or~\circnum{13} and~\circnum{14}), we can clearly see that the non-linear ranking aggregation successfully boosts the ranking consistency w.r.t the performance.
\begin{figure}[t]
  \small
  \begin{center}
     \includegraphics[keepaspectratio=true,width=0.6\columnwidth]{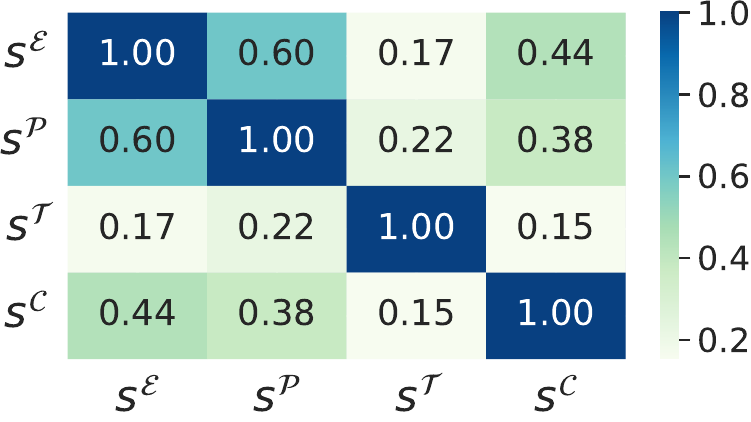}
  \end{center}
    \vspace{-0.5cm}
     \caption{Correlation analysis on the zero-cost proxies of AZ-NAS. We report Kendall's $\tau$ between the estimated network rankings on ImageNet16-120 of NAS-Bench-201~\cite{dong2020nasbench201}.}
    \vspace{-0.6cm}
  \label{fig:confusion_matrix}
\end{figure}
The aggregation is particularly useful with more proxies, verifying its significance in AZ-NAS. (3)~The trainability proxy alone might not be effective as shown in \circnum{3}, since it mainly focuses on the stable gradient propagation, without considering \eg, a network's capacity to learn. Nevertheless, we can see from \circnum{5} to \circnum{11} that the trainability proxy helps others to achieve strong ranking consistency, especially on ImageNet16-120. For an in-depth analysis, we show in Fig.~\ref{fig:confusion_matrix} how the proposed proxies are correlated with each other on ImageNet16-120. We can see that the trainability proxy is less correlated with others. Based on these observations, we can conclude that coupling less correlated proxies improves the NAS performance more significantly~(\eg, the results on ImageNet16-120 in~\circnum{6} and~\circnum{8}). This is because such proxies tend to capture unique network characteristics, making them mutually complementary and leading to a synergy effect. AZ-NAS fully leverages the complementary features among the proxies, boosting the NAS performance drastically.

\vspace{-0.35cm}
\paragraph{Assembling other zero-cost proxies.}
AZ-NAS is built upon the idea that assembling various zero-cost proxies, similar to ensemble learning, could bring better performance for training-free NAS. To further verify this idea, we incorporate the proposed proxies into existing ones~\cite{abdelfattah2021zerocost,tanaka2020pruning,li2023zico} and show in Table~\ref{tab:ensemble_others} the results on NAS-Bench-201~\cite{dong2020nasbench201}. We can see that an ensemble of our zero-cost proxies and the previous one improves the ranking consistency w.r.t the performance consistently, suggesting that the basic idea of AZ-NAS is also applicable to other proxies. Similar to the observation in the ablation study, we can obtain better ranking consistency by exploiting more proxies. In particular, we can further boost the performance in terms of Kendall's~$\tau$ by aggregating all of our proxies and the previous one, at the cost of the additional runtime.

\begin{table}[t]
  \small
  \setlength{\tabcolsep}{0.15cm}
  \centering
  \caption{Quantitative comparison of incorporating our zero-cost proxies with existing ones. We adopt our non-linear ranking aggregation method to estimate the predicted network rankings. We report the correlation coefficients~(Kendall's~$\tau$) between predicted and ground-truth network rankings on NAS-Bench-201~\cite{dong2020nasbench201}, together with average runtimes.}
  \vspace{-0.3cm}
  \begin{adjustbox}{max width=\columnwidth}
     \begin{tabular}{l c c c c}
        \toprule
        \multirow{2}{*}{Aggregated zero-cost proxies}    & \multirow{2}{*}{CIFAR-10} & \multirow{2}{*}{CIFAR-100} & \multirow{2}{*}{IN16-120} & \multirow{2}{*}{\shortstack[c]{Runtime \\ (ms/arch)}} \\ \\
        \midrule
        $s^{\mathcal{C}}$                                                             & 0.578    & 0.551     & 0.517 & - \\
        $s^{\mathcal{C}}$ + $s^{\mathcal{E}}$                                         & 0.674    & 0.653     & 0.601 & 22.3 \\
        $s^{\mathcal{C}}$ + $s^{\mathcal{E}}$ + $s^{\mathcal{T}}$                     & 0.731    & 0.714     & 0.708 & 42.7 \\
        $s^{\mathcal{C}}$ + $s^{\mathcal{E}}$ + $s^{\mathcal{T}}$ + $s^{\mathcal{P}}$ & 0.741    & 0.723     & 0.710 & 42.7 \\
        \midrule
        ZiCo~\cite{li2023zico}                                                               & 0.589    & 0.590     & 0.584 & 372.8 \\
        ZiCo + $s^{\mathcal{C}}$                                                             & 0.632    & 0.615     & 0.595 & 372.8 \\
        ZiCo + $s^{\mathcal{C}}$ + $s^{\mathcal{E}}$                                         & 0.733    & 0.717     & 0.678 & 395.1 \\
        ZiCo + $s^{\mathcal{C}}$ + $s^{\mathcal{E}}$ + $s^{\mathcal{T}}$                     & 0.761    & 0.749     & 0.743 & 415.5 \\
        ZiCo + $s^{\mathcal{C}}$ + $s^{\mathcal{E}}$ + $s^{\mathcal{T}}$ + $s^{\mathcal{P}}$ & 0.773    & 0.757     & 0.747 & 415.5 \\
        \midrule
        Synflow~\cite{abdelfattah2021zerocost,tanaka2020pruning}                                & 0.571    & 0.565     & 0.555 & 53.4 \\
        Synflow + $s^{\mathcal{C}}$                                                             & 0.636    & 0.616     & 0.594 & 53.4 \\
        Synflow + $s^{\mathcal{C}}$ + $s^{\mathcal{E}}$                                         & 0.741    & 0.721     & 0.681 & 75.7 \\
        Synflow + $s^{\mathcal{C}}$ + $s^{\mathcal{E}}$ + $s^{\mathcal{T}}$                     & 0.768    & 0.753     & 0.746 & 96.1 \\
        Synflow + $s^{\mathcal{C}}$ + $s^{\mathcal{E}}$ + $s^{\mathcal{T}}$ + $s^{\mathcal{P}}$ & 0.776    & 0.758     & 0.747 & 96.1 \\
        \bottomrule
     \end{tabular} \label{tab:ensemble_others}
  \end{adjustbox}
  \vspace{-0.35cm}
\end{table}

\section{Conclusion} \label{sec:conclusion}
We have presented AZ-NAS, a training-free NAS approach that assembles various zero-cost proxies to substantially enhance the NAS performance. To this end, we have designed novel zero-cost proxies from distinct and complementary perspectives. We have also proposed to integrate them into a final score effectively with a non-linear ranking aggregation technique. Extensive experiments clearly demonstrate the efficiency and efficacy of AZ-NAS, surpassing previous training-free NAS methods with a reasonably fast runtime. We expect that AZ-NAS can provide a plug-and-play solution, allowing other NAS methods to maximize the performance by seamlessly incorporating our zero-cost proxies during the search, with a minimal increase in the computational cost.

\vspace{-0.3cm}
\paragraph{Acknowledgments.}
This work was partly supported by IITP grants funded by the Korea government (MSIT) (No.RS-2022-00143524, Development of Fundamental Technology and Integrated Solution for Next-Generation Automatic Artificial Intelligence System, No.2022-0-00124, Development of Artificial Intelligence Technology for Self-Improving Competency-Aware Learning Capabilities) and the KIST Institutional Program (Project No.2E31051-21-203).


\clearpage
{
    \small
    \bibliographystyle{ieeenat_fullname}
    \bibliography{main}
}

\clearpage

\maketitlesupplementary

In this supplement, we provide additional results and in-depth analyses of AZ-NAS on the NDS~\cite{radosavovic2019network}, NAS-Bench-201~\cite{dong2020nasbench201}, and MobileNetV2~\cite{sandler2018mobilenetv2,lin2021zen} search spaces.

\vspace{-0.2cm}
\paragraph{Additional results on the NDS benchmark.}
The Neural Design Space~(NDS)~\cite{radosavovic2019network} benchmark provides the ground-truth accuracies of candidate architectures for several cell-based search spaces, where each space adopts a distinct set of candidate operations. We present in Table~\ref{tab:NDS} the quantitative comparison of AZ-NAS with the state of the art~\cite{tanaka2020pruning,mellor2021neural,li2023zico,abdelfattah2021zerocost} on the NDS benchmark. For each search space in NDS, we report a correlation coefficient~(Kendall's~$\tau$) between predicted and ground-truth network rankings, and an average top-1 accuracy of selected networks on CIFAR-10~\cite{krizhevsky2009learning}. We can see that AZ-NAS shows superior results across the search spaces, verifying the generalization ability of AZ-NAS on various search spaces. On the other hand, other methods~\cite{tanaka2020pruning,mellor2021neural,li2023zico,abdelfattah2021zerocost} exploiting a single proxy are either outperformed by or merely comparable to \#Params or FLOPs in terms of ranking consistency w.r.t the performance~(\ie, Kendall's~$\tau$), indicating that they are less effective on the NDS benchmark than NAS-Bench-201~\cite{dong2020nasbench201}~(see Table 1 in the main paper). These results confirm that assembling various proxies could improve the robustness against the variability of search spaces.

\vspace{-0.2cm}
\paragraph{Reproducibility of training-free NAS methods.}
In Table~2 of the main paper, we compare AZ-NAS with the state of the art, mainly focusing on the final performance of selected networks. Considering the randomness during the search phase, \eg, due to a sampling process of candidate architectures from a large-scale search space~\cite{sandler2018mobilenetv2,lin2021zen}, it is also crucial to achieve consistent NAS results over multiple trials for practical use. We thus evaluate the reproducibility of AZ-NAS and the state-of-the-art training-free NAS methods~\cite{lin2021zen,li2023zico} under a fair experimental setting. Specifically, for each method, we find three network architectures from scratch with different seed numbers on the MobileNetV2 search space~\cite{sandler2018mobilenetv2,lin2021zen}. We use the same number of search iterations~(100K) and FLOPs constraint~(450M) for all the methods for a fair comparison. We train selected networks on ImageNet~\cite{deng2009imagenet} with a simplified training scheme used in~\cite{li2023zico} to reduce the training cost. Compared to the training setting for the experiments in Table~2 of the main paper, the simplified one reduces training epochs from 480 to 150, while not incorporating the teacher-student distillation~\cite{hinton2015distilling} and advanced data augmentation techniques~(\eg, AutoAugment~\cite{cubuk2019autoaugment}, MixUp~\cite{zhang2018mixup}, and RandomErase~\cite{zhong2020random}).

We present in Table~\ref{tab:MBV2_reproduce} the average and standard deviation values of top-1 validation accuracies on ImageNet for selected networks. While the network architectures provided by the authors of ZenNAS~\cite{lin2021zen} and ZiCo~\cite{li2023zico} show decent performance, we could not reproduce comparable results within three random runs, suggesting that these methods lack reproducibility. On the contrary, AZ-NAS outperforms the others by significant margins in terms of the average accuracy, while showing the lowest deviation, demonstrating its ability to find high-performing networks consistently across multiple runs. We can also see that the relative order of the search costs, measured under a fair search configuration using the same machine, is aligned with the one in Table 1 of the main paper, and AZ-NAS offers a good compromise between the NAS performance and search cost.

\begin{table}[t]
  \small
  \setlength{\tabcolsep}{0.12cm}
  \centering
  \caption{Quantitative comparison on the NDS~\cite{radosavovic2019network} benchmark. For each search space, we report Kendall's $\tau$~(KT) using all candidate architectures. We also present average and standard deviation of test accuracies~(Acc.) for selected networks on CIFAR-10~\cite{krizhevsky2009learning}, which are obtained over 5 random runs. To this end, we randomly sample 1000 candidate architectures for each run and select an optimal one among them based on each training-free NAS method.}
  \vspace{-0.25cm}
  \begin{adjustbox}{max width=\columnwidth}
    \begin{tabular}{l c c c c c c}
      \toprule
      \multirow{2}{*}[-2pt]{Method} & \multicolumn{2}{c}{ENAS} & \multicolumn{2}{c}{Amoeba} & \multicolumn{2}{c}{NASNet} \\ \cmidrule(lr){2-3} \cmidrule(lr){4-5} \cmidrule(lr){6-7}
               & KT    & Acc.                           & KT        & Acc.                       & KT        & Acc. \\ \midrule 
      \#Params & 0.411 & 92.94 $\pm$ 1.26               & 0.241     & 76.77 $\pm$ 37.3           & 0.289     & 92.56 $\pm$ 1.10 \\ 
      FLOPs    & 0.409 & 92.94 $\pm$ 1.26               & 0.239     & 76.77 $\pm$ 37.3           & 0.276     & 92.56 $\pm$ 1.10 \\ 
      Synflow~\cite{tanaka2020pruning,abdelfattah2021zerocost}  & 0.116 & 67.90 $\pm$ 28.9               & -0.076    & 87.70 $\pm$ 8.29           & 0.007     & 79.98 $\pm$ 25.0 \\ 
      NASWOT~\cite{mellor2021neural}   & 0.375 & 93.56 $\pm$ 1.55               & 0.192     & 92.89 $\pm$ 0.29           & 0.286     & 93.55 $\pm$ 0.93 \\ 
      ZiCo~\cite{li2023zico}     & 0.200 & 92.19 $\pm$ 1.15               & -0.016    & 92.25 $\pm$ 0.60           & 0.089     & 92.72 $\pm$ 0.88 \\ 
      AZ-NAS   & \B{0.495} & {\B{94.41}} $\pm$ \B{0.13} & \B{0.386} & {\B{93.75}} $\pm$ \B{0.24} & \B{0.426} & {\B{93.72}} $\pm$ \B{0.50} \\
      \bottomrule
    \end{tabular} \label{tab:NDS}
  \end{adjustbox}
  \vspace{-0.2cm}
\end{table}

\begin{table}[t]
  \small
  \setlength{\tabcolsep}{0.15cm}
  \centering
  \caption{Comparison of the reproducibility for the state-of-the-art training-free NAS methods~\cite{lin2021zen,li2023zico} and ours on the MobileNetV2~\cite{sandler2018mobilenetv2} search space~\cite{lin2021zen}. For each method, we report the average and standard deviation values of top-1 validation accuracies on ImageNet~\cite{deng2009imagenet}, obtained over three random runs \textit{starting from the search phase}, together with the search costs in terms of GPU hours. The results of~\cite{lin2021zen,li2023zico} are obtained using the official code provided by the authors.} 
  \vspace{-0.3cm}
  \begin{adjustbox}{max width=\columnwidth}
     \begin{tabular}{l c c c}
        \multicolumn{4}{r}{\textsuperscript{$\dagger$}\footnotesize{Architectures provided by the authors.}}\\
        \multicolumn{4}{r}{\textsuperscript{*}\footnotesize{An architecture found with a different FLOPs budget~(\ie, 400M).}}\\
        \toprule
        \multirow{2}{*}{Method} & \multirow{2}{*}{FLOPs} & \multirow{2}{*}{Top-1 acc.} & \multirow{2}{*}{\shortstack[c]{Search cost \\ (GPU hours)}} \\
        \\
        \midrule
        ZenNAS\textsuperscript{$\dagger$}~\cite{lin2021zen} & 410M\textsuperscript{*} & 75.87                        & - \\
        ZiCo\textsuperscript{$\dagger$}~\cite{li2023zico}   & 448M                    & 76.07                        & - \\
        \midrule
        ZenNAS~\cite{lin2021zen}                            & 458M $\pm$ 1.6M         & 73.76 $\pm$ 1.32             & 4.9 \\
        ZiCo~\cite{li2023zico}                              & 450M $\pm$ 5.0M         & 72.46 $\pm$ 0.84             & 15.7 \\
        AZ-NAS~(Ours)                                       & 462M $\pm$ 1.5M         & {\B{76.46}} $\pm$ {\B{0.06}} & 10.0 \\
        \bottomrule
     \end{tabular} \label{tab:MBV2_reproduce}
  \end{adjustbox}
  \vspace{-0.25cm}
\end{table}

\begin{figure*}[t]
  \small
  \begin{center}
    \begin{subfigure}{0.23\textwidth}
      \centering
      \includegraphics[width=1\textwidth]{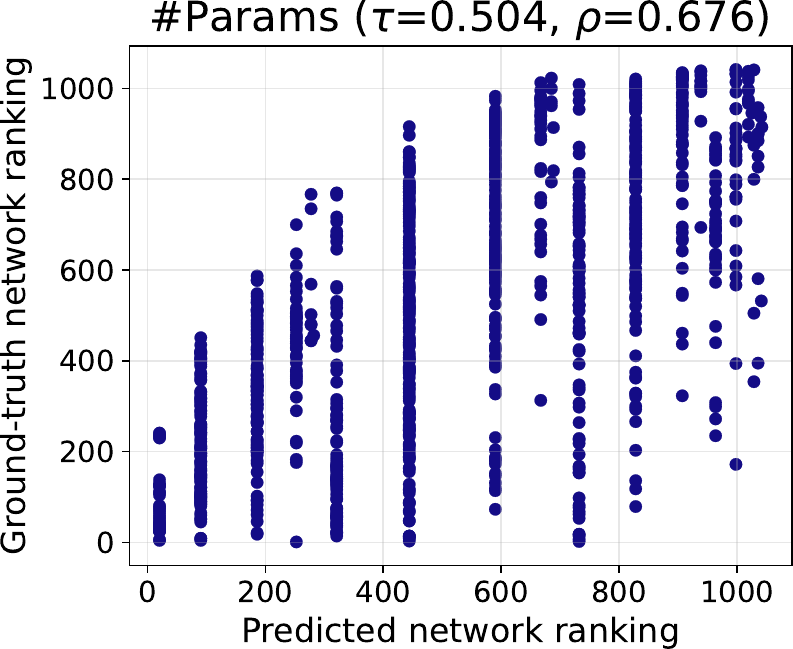}
      \caption{\#Params.}
    \end{subfigure}
    \hfill
    \begin{subfigure}{0.23\textwidth}
      \centering
      \includegraphics[width=1\textwidth]{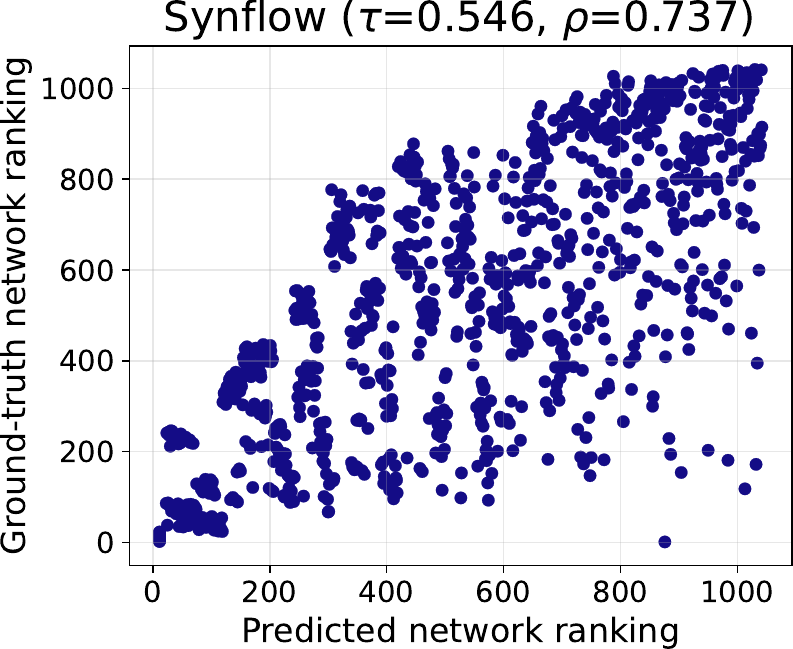}
      \caption{Synflow~\cite{tanaka2020pruning}.}
    \end{subfigure}
    \hfill
    \begin{subfigure}{0.23\textwidth}
      \centering
      \includegraphics[width=1\textwidth]{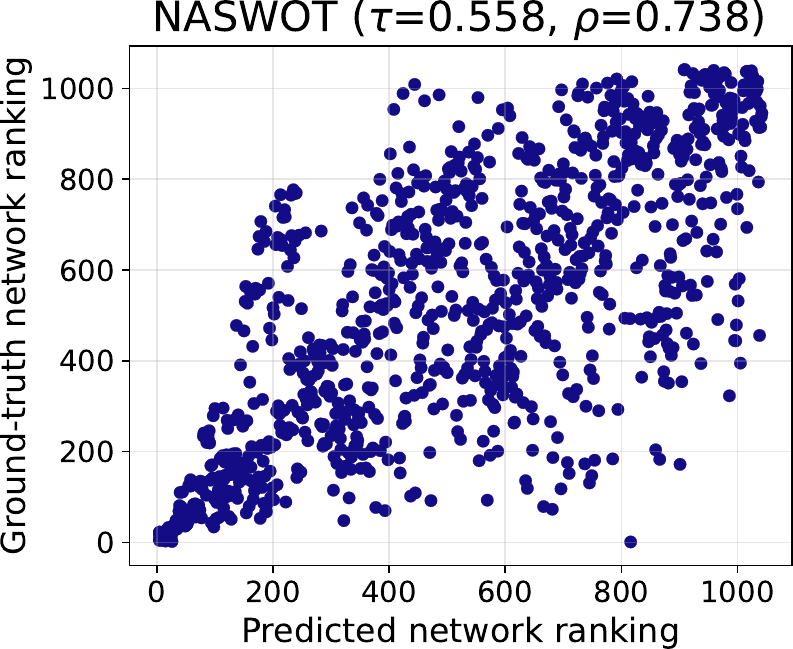}
      \caption{NASWOT~\cite{mellor2021neural}.}
    \end{subfigure}
    \hfill
    \begin{subfigure}{0.23\textwidth}
      \centering
      \includegraphics[width=1\textwidth]{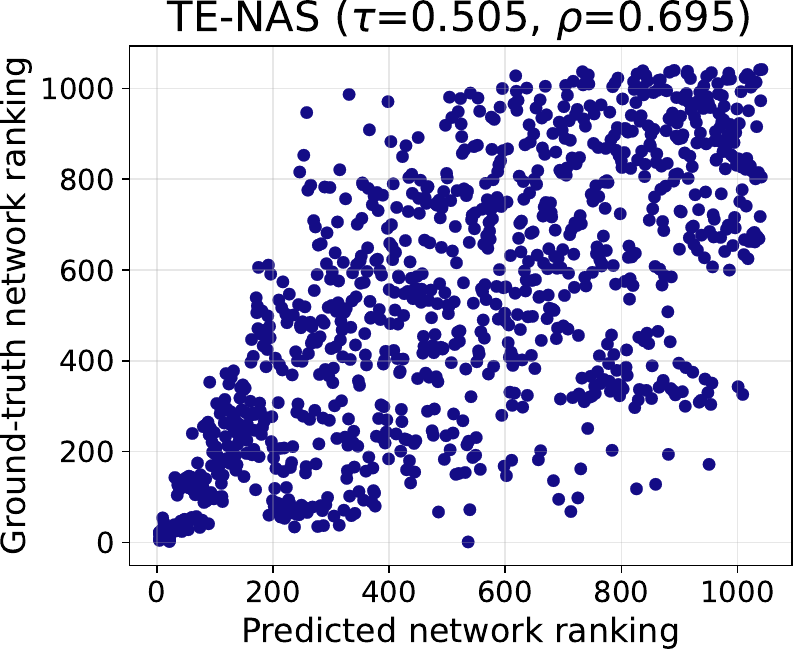}
      \caption{TE-NAS~\cite{chen2020tenas}.}
    \end{subfigure}

    \vspace{0.2cm}

    \begin{subfigure}{0.23\textwidth}
      \centering
      \includegraphics[width=1\textwidth]{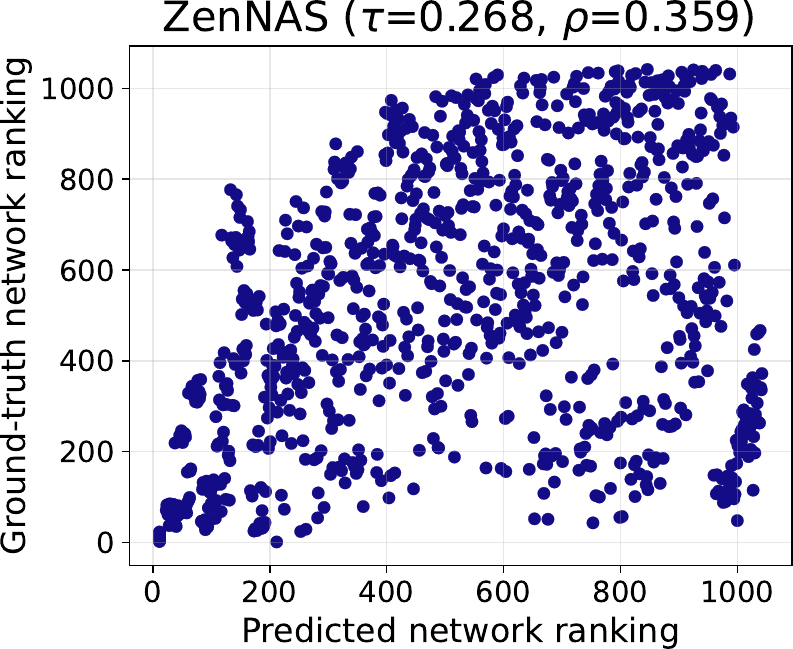}
      \caption{ZenNAS~\cite{lin2021zen}.}
    \end{subfigure}
    \hfill
    \begin{subfigure}{0.23\textwidth}
      \centering
      \includegraphics[width=1\textwidth]{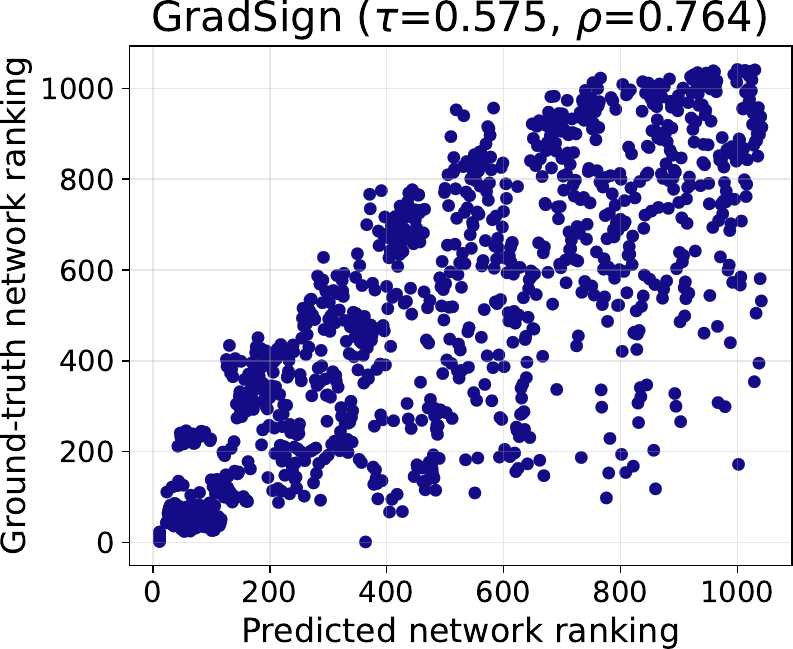}
      \caption{GradSign~\cite{zhang2022gradsign}.}
    \end{subfigure}
    \hfill
    \begin{subfigure}{0.23\textwidth}
      \centering
      \includegraphics[width=1\textwidth]{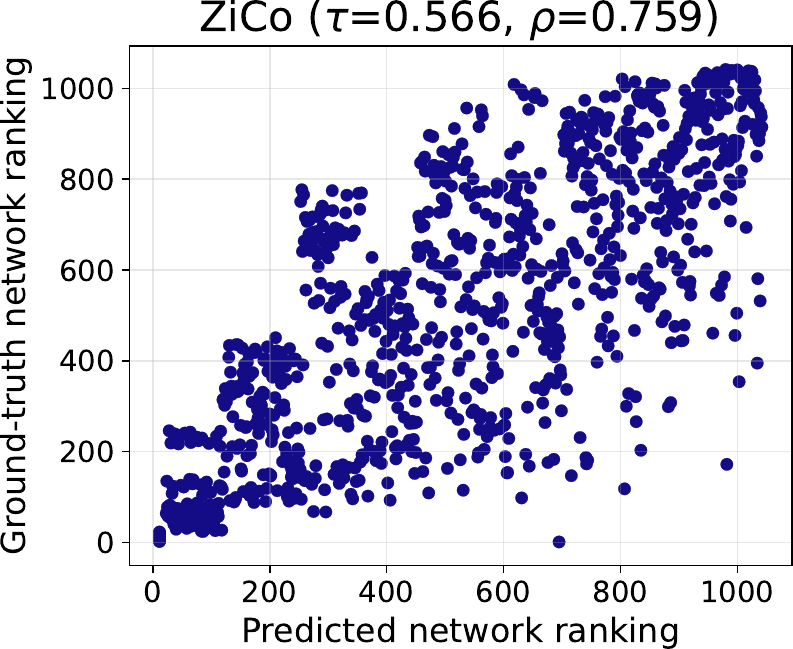}
      \caption{ZiCo~\cite{li2023zico}.}
    \end{subfigure}
    \hfill
    \begin{subfigure}{0.23\textwidth}
      \centering
      \includegraphics[width=1\textwidth]{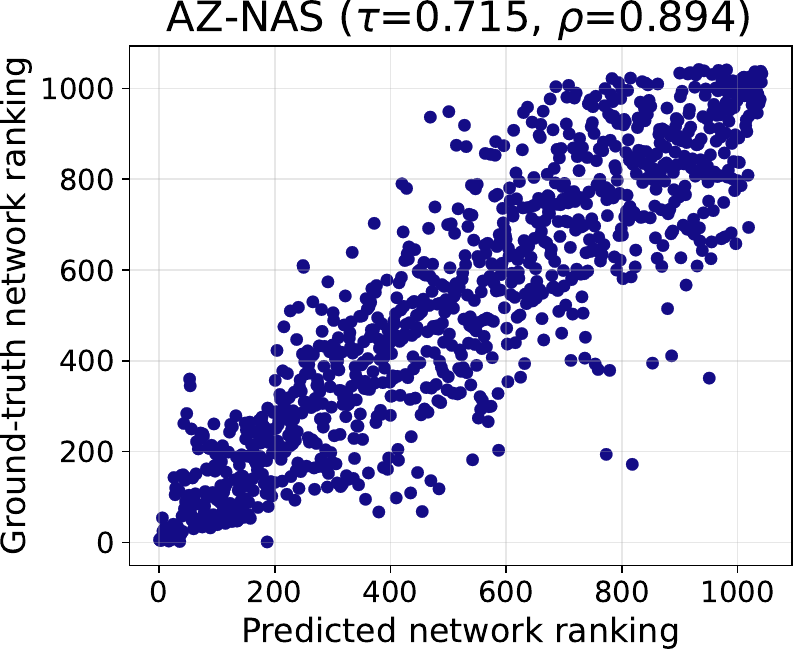}
      \caption{AZ-NAS (Ours).}
    \end{subfigure}
  \end{center}
    \vspace{-0.5cm}
     \caption{Visual comparison of training-free NAS methods in terms of predicted network ranking~($x$-axis) \vs ground truth~($y$-axis) on ImageNet16-120 of NAS-Bench-201~\cite{dong2020nasbench201}. We report the correlation coefficients between them in terms of Kendall's~$\tau$ and Spearman's~$\rho$, denoted by~$\tau$ and~$\rho$, respectively.}
  \label{fig:scatter_sota}
  \vspace{0.5cm}
\end{figure*}

\begin{figure*}[t]
  \small
  \begin{center}
    \begin{subfigure}{0.3\textwidth}
      \centering
      \includegraphics[width=1\textwidth]{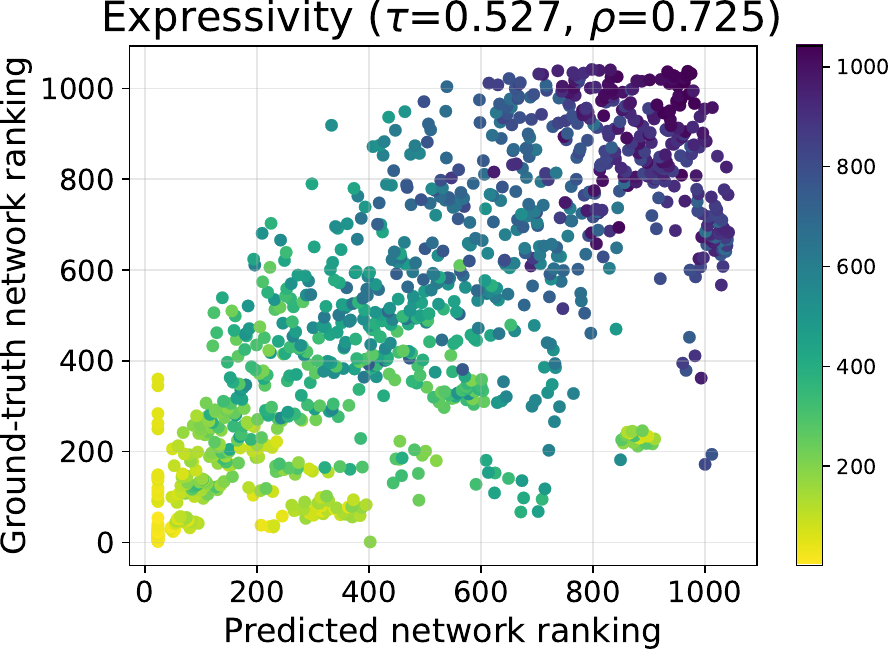}
      \caption{$s^{\mathcal{E}}$.}
    \end{subfigure}
    \hfill
    \begin{subfigure}{0.3\textwidth}
      \centering
      \includegraphics[width=1\textwidth]{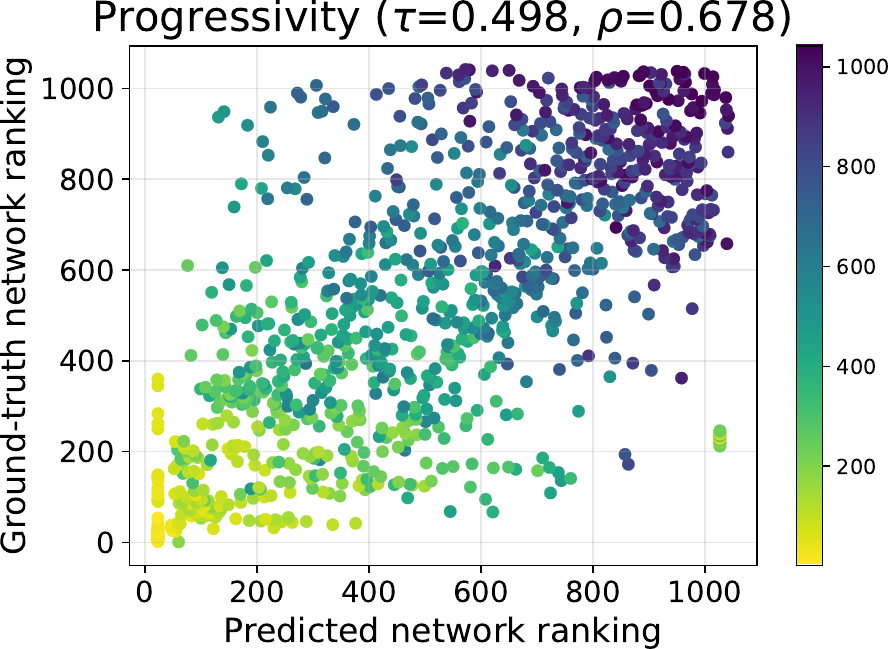}
      \caption{$s^{\mathcal{P}}$.}
    \end{subfigure}
    \hfill
    \begin{subfigure}{0.3\textwidth}
      \centering
      \includegraphics[width=1\textwidth]{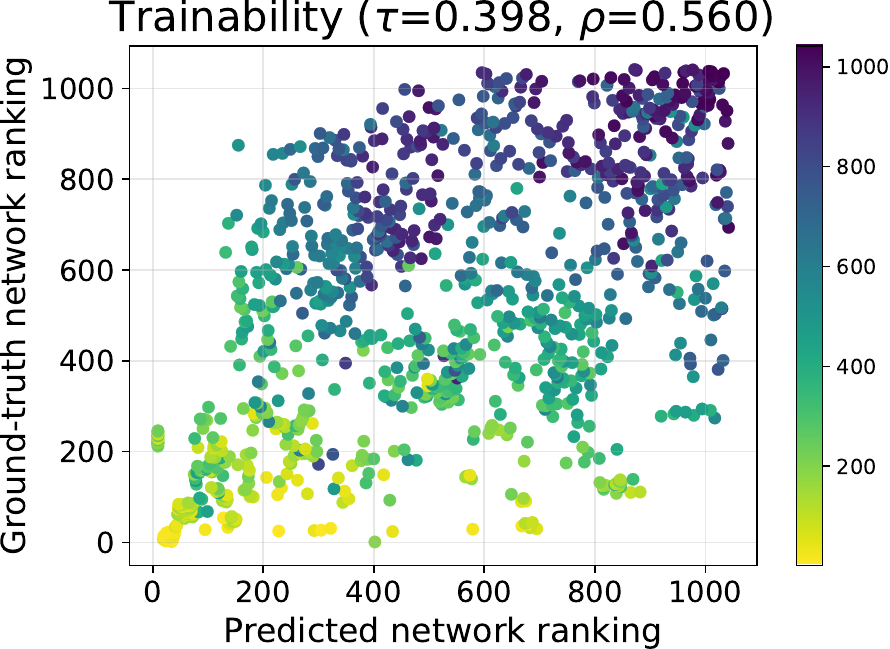}
      \caption{$s^{\mathcal{T}}$.}
    \end{subfigure}

    \vspace{0.2cm}

    \begin{subfigure}{0.3\textwidth}
      \centering
      \includegraphics[width=1\textwidth]{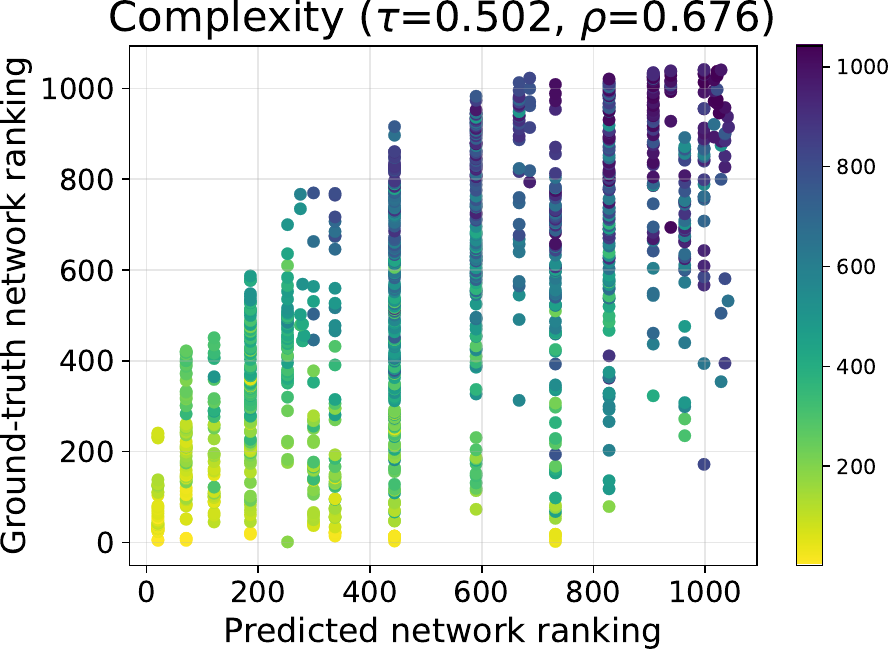}
      \caption{$s^{\mathcal{C}}$.}
    \end{subfigure}
    \hfill
    \begin{subfigure}{0.3\textwidth}
      \centering
      \includegraphics[width=1\textwidth]{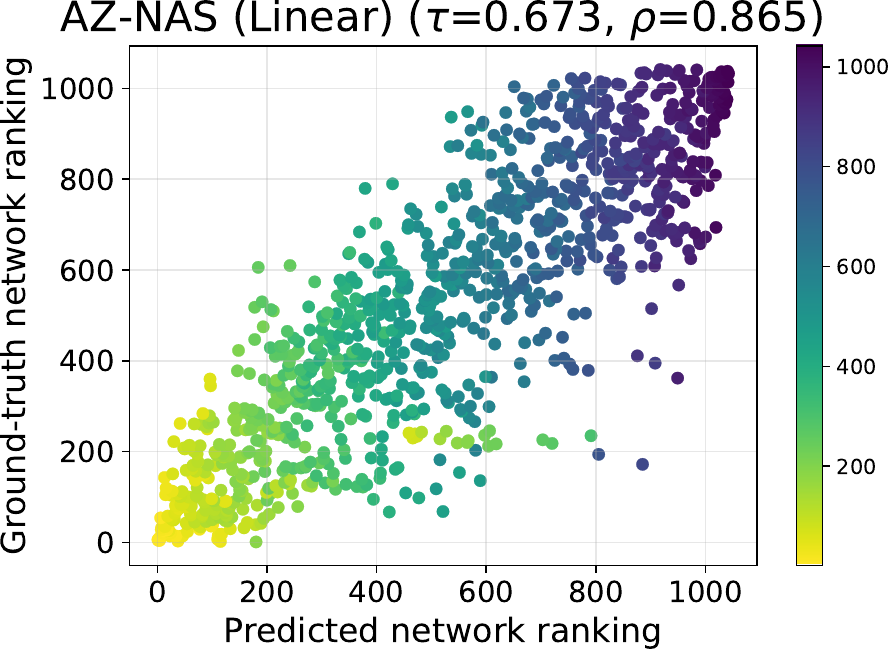}
      \caption{$s^{\textrm{AZ}}$~(Linear ranking aggregation).}
    \end{subfigure}
    \hfill
    \begin{subfigure}{0.3\textwidth}
      \centering
      \includegraphics[width=1\textwidth]{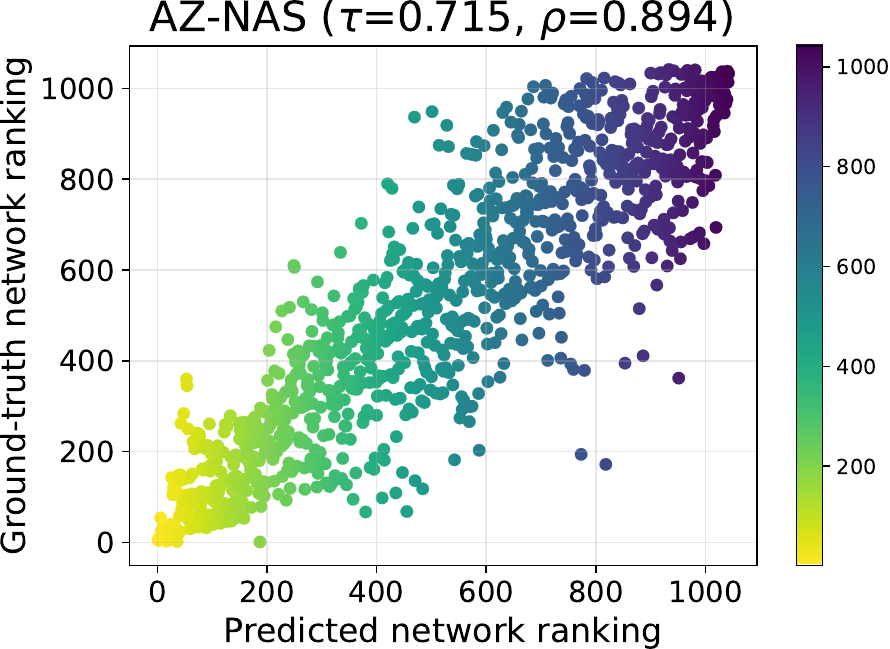}
      \caption{$s^{\textrm{AZ}}$~(Non-linear ranking aggregation).}
    \end{subfigure}
  \end{center}
  \vspace{-0.5cm}
     \caption{Visual comparison of the zero-cost proxies of AZ-NAS~((a)-(d)), and the linear and non-linear ranking aggregation methods~((e) and (f)), in terms of predicted network ranking~($x$-axis) \vs ground truth~($y$-axis) on ImageNet16-120 of NAS-Bench-201~\cite{dong2020nasbench201}. The colors of the points, ranging from light-yellow to dark-blue, correspond to the network ranking in (f) predicted by the AZ-NAS score. (Best viewed in color.)}
  \label{fig:scatter_ours}
\end{figure*}

\vspace{-0.2cm}
\paragraph{Visualization of ranking consistency.}
We present in Fig.~\ref{fig:scatter_sota} a visual comparison between the training-free NAS methods~\cite{tanaka2020pruning,mellor2021neural,chen2020tenas,lin2021zen,zhang2022gradsign,li2023zico} and ours, where each plot shows the predicted network ranking~($x$-axis) \vs the ground-truth network ranking~($y$-axis) on ImageNet16-120 of NAS-Bench-201~\cite{dong2020nasbench201}. For visualization, we exploit the same 1042 candidate architectures for all the methods, evenly sampled according to the test accuracy on ImageNet16-120. From Figs.~\ref{fig:scatter_sota}(a)-(g), we can see that previous training-free NAS methods produce incorrect predictions frequently, such as highly-ranked networks with low ground-truth performance that can be found in the bottom-right regions of the plots, making it difficult to discover a high-performing network accurately. On the other hand, the network ranking predicted by AZ-NAS in Fig.~\ref{fig:scatter_sota}(h) shows the strongest correlation with the ground truth, with the points concentrated closely around the $y=x$ line in the plot, demonstrating the superiority of our method. 

We also compare in Fig.~\ref{fig:scatter_ours} the network rankings obtained with each of the zero-cost proxies of AZ-NAS, and the aggregated ones using either linear or non-linear methods on ImageNet16-120 of NAS-Bench-201. We can observe in Figs.~\ref{fig:scatter_ours}(a)-(d) that the network rankings estimated by a single proxy alone show weak consistency w.r.t the ground truth. For example, the proxies in Figs.~\ref{fig:scatter_ours}(a)-(d) often assign high scores for low-performing networks~(\eg, bottom-right regions of Figs.~\ref{fig:scatter_ours}(c) and (d)) or low scores for high-performing networks~(\eg, top-left regions of Figs.~\ref{fig:scatter_ours}(b) and (c)). By combining the proxies into the AZ-NAS score in Fig.~\ref{fig:scatter_ours}(f), we can improve the ranking consistency substantially. We can find that the yellow-green and green-blue points near the bottom-right and top-left regions in Figs.~\ref{fig:scatter_ours}(a)-(d) are located closely to the~$y=x$ line in Fig.~\ref{fig:scatter_ours}(f), suggesting that the misaligned rankings estimated by individual proxies are corrected by assembling. By comparing the results obtained with the linear and non-linear ranking aggregation methods in Figs.~\ref{fig:scatter_ours}(e) and~(f), respectively, we can also see that the non-linear one makes the ranking consistency much stronger, while correcting erroneous predictions~(see yellow-green points in the bottom-right region of Fig.~\ref{fig:scatter_ours}(e)), clearly demonstrating the effectiveness of our non-linear ranking aggregation method.

\begin{table}[t]
  \small
  \setlength{\tabcolsep}{0.15cm}
  \centering
  \caption{Comparison of AZ-NAS using different network initialization methods. We report correlation coefficients~(Kendall's $\tau$) between predicted and ground-truth network rankings on NAS-Bench-201~\cite{dong2020nasbench201}. For other experiments using convolutional neural networks, we adopt the Kaiming normal initialization with a fan-in mode by default (highlighted by a gray color).}
  \vspace{-0.3cm}
  \begin{adjustbox}{max width=\columnwidth}
     \begin{tabular}{L{3.4cm} C{1.5cm} C{1.5cm} C{1.5cm}}
        \toprule
        Initialization method    & CIFAR-10 & CIFAR-100 & IN16-120 \\ \rowcolor{Gainsboro!60}
        \midrule
        Kaiming normal (fan-in)      & 0.741    & 0.723     & 0.710 \\
        Kaiming normal (fan-out)     & 0.754    & 0.733     & 0.724 \\
        Xavier normal                & 0.754    & 0.732     & 0.724 \\
        Normal (std=0.1)             & 0.740    & 0.715     & 0.709 \\
        Uniform ([-0.1, 0.1])        & 0.730    & 0.703     & 0.702 \\
        \bottomrule
     \end{tabular} \label{tab:init_methods}
  \end{adjustbox}
  \vspace{-0.15cm}
\end{table}

\begin{table}[t]
  \small
  \setlength{\tabcolsep}{0.15cm}
  \centering
  \caption{Comparison of AZ-NAS using different batch sizes in the search phase. We report the correlation coefficients~(Kendall's~$\tau$) between predicted and ground-truth network rankings on NAS-Bench-201~\cite{dong2020nasbench201}. For other experiments, we adopt the batch size of 64 by default (highlighted by a gray color).}
  \vspace{-0.3cm}
  \begin{adjustbox}{max width=\columnwidth}
     \begin{tabular}{L{1.4cm} C{1.5cm} C{1.5cm} C{1.5cm} C{1.5cm}}
        \toprule
        \multirow{2}{*}{Batch size} & \multirow{2}{*}{CIFAR-10} & \multirow{2}{*}{CIFAR-100} & \multirow{2}{*}{IN16-120} & \multirow{2}{*}{\shortstack[c]{Runtime\\(ms/arch)}} \\ \\
        \midrule
        8          & 0.730    & 0.709     & 0.687    & 37.33 \\
        16         & 0.736    & 0.717     & 0.701    & 39.39 \\
        32         & 0.740    & 0.721     & 0.708    & 39.78 \\ \rowcolor{Gainsboro!60}
        64         & 0.741    & 0.723     & 0.710    & 42.71 \\
        128        & 0.741    & 0.724     & 0.710    & 54.32 \\
        \bottomrule
     \end{tabular} \label{tab:batch_sizes}
  \end{adjustbox}
  \vspace{-0.25cm}
\end{table}

\vspace{-0.2cm}
\paragraph{Robustness to various initialization methods.}
To verify the robustness of AZ-NAS against network initialization methods, we compare the NAS performance by initializing weight parameters from Kaiming normal~\cite{he2015delving}, Xavier normal~\cite{glorot2010understanding}, normal, or uniform distributions. We provide the results in Table~\ref{tab:init_methods} in terms of the ranking consistency~(Kendall's $\tau$) w.r.t the ground-truth performance on NAS-Bench-201~\cite{dong2020nasbench201}. We can see that AZ-NAS shows satisfactory NAS results consistently, regardless of the types of initialization methods. Note that we simply adopt Kaiming normal with a fan-in mode implemented in PyTorch~\cite{paszke2019pytorch} as our default initialization method for the experiments using convolutional neural networks. We can further improve the NAS performance when exploiting other initialization methods, such as Kaiming normal with a fan-out mode or Xavier normal.

\vspace{-0.25cm}
\paragraph{Comparison using various batch sizes for search.}
We present in Table~\ref{tab:batch_sizes} the ranking consistency~(Kendall's $\tau$) of AZ-NAS w.r.t the performance on NAS-Bench-201~\cite{dong2020nasbench201} using different batch sizes in the search phase, in order to understand how AZ-NAS is affected by a batch size. We can see that AZ-NAS achieves good NAS performance even when a small batch size (\ie, 8 in the first row) is used for search. As we increase the batch size, the ranking consistency gradually improves at the expense of additional runtime. We can observe that the batch size of 64 provides the best trade-off between runtime and ranking consistency, and we adopt this as a default setting for other experiments.

\begin{figure*}[t]
  \small
  \begin{center}
    \centering
    \includegraphics[width=0.78\textwidth]{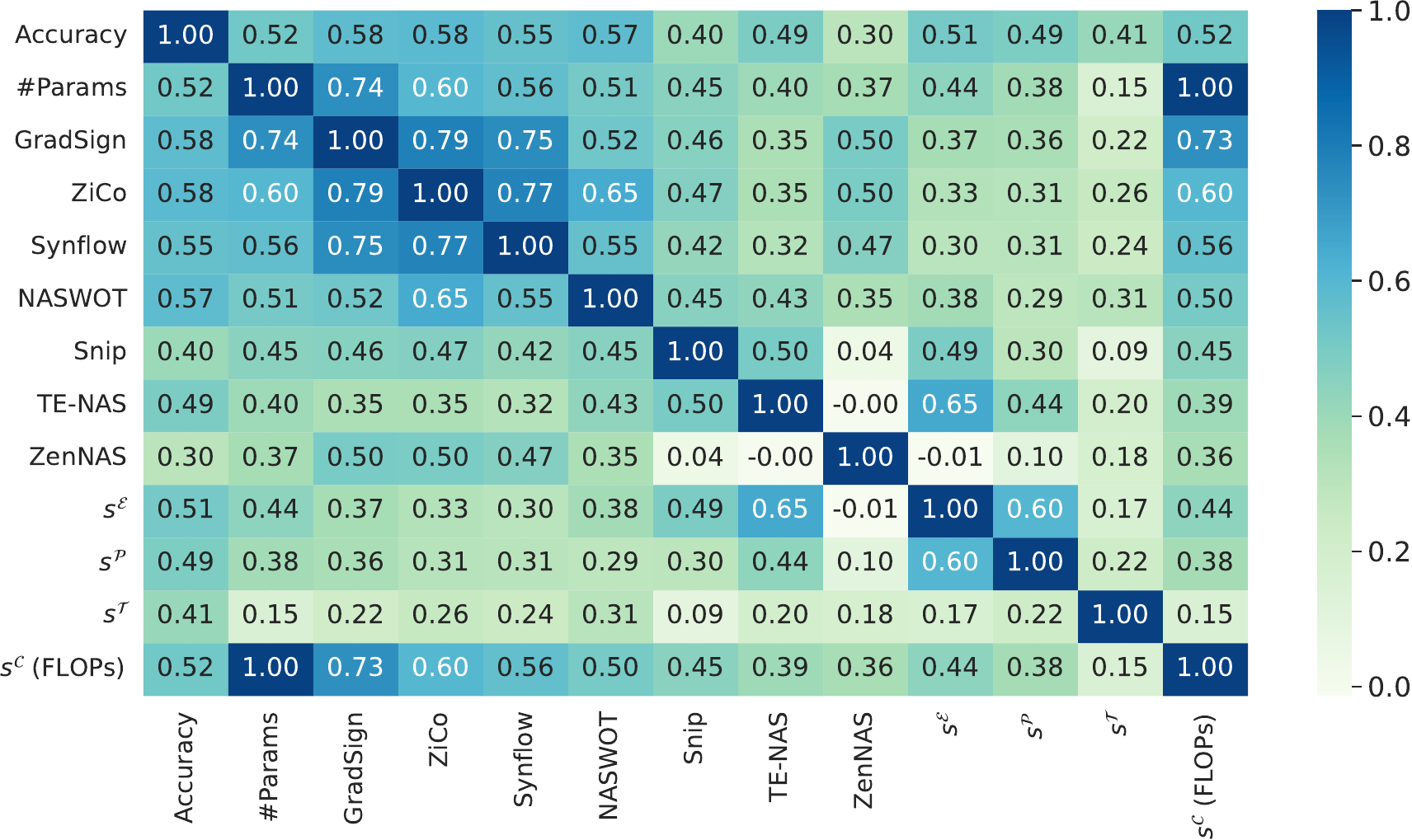}  
    \vspace{-0.1cm}
    \caption{Correlation analysis of various zero-cost proxies and ours on ImageNet16-120 of NAS-Bench-201~\cite{dong2020nasbench201}. We report correlation coefficients~(Kendall's~$\tau$) for the pairs of two network rankings estimated by the zero-cost proxies in the $x$- and $y$-axes, respectively.}
    \label{fig:intercorrelation}
  \end{center}
  \vspace{-0.2cm}
\end{figure*}

\begin{figure*}[t]
  \small
  \begin{center}
    \centering
  \includegraphics[width=0.78\textwidth]{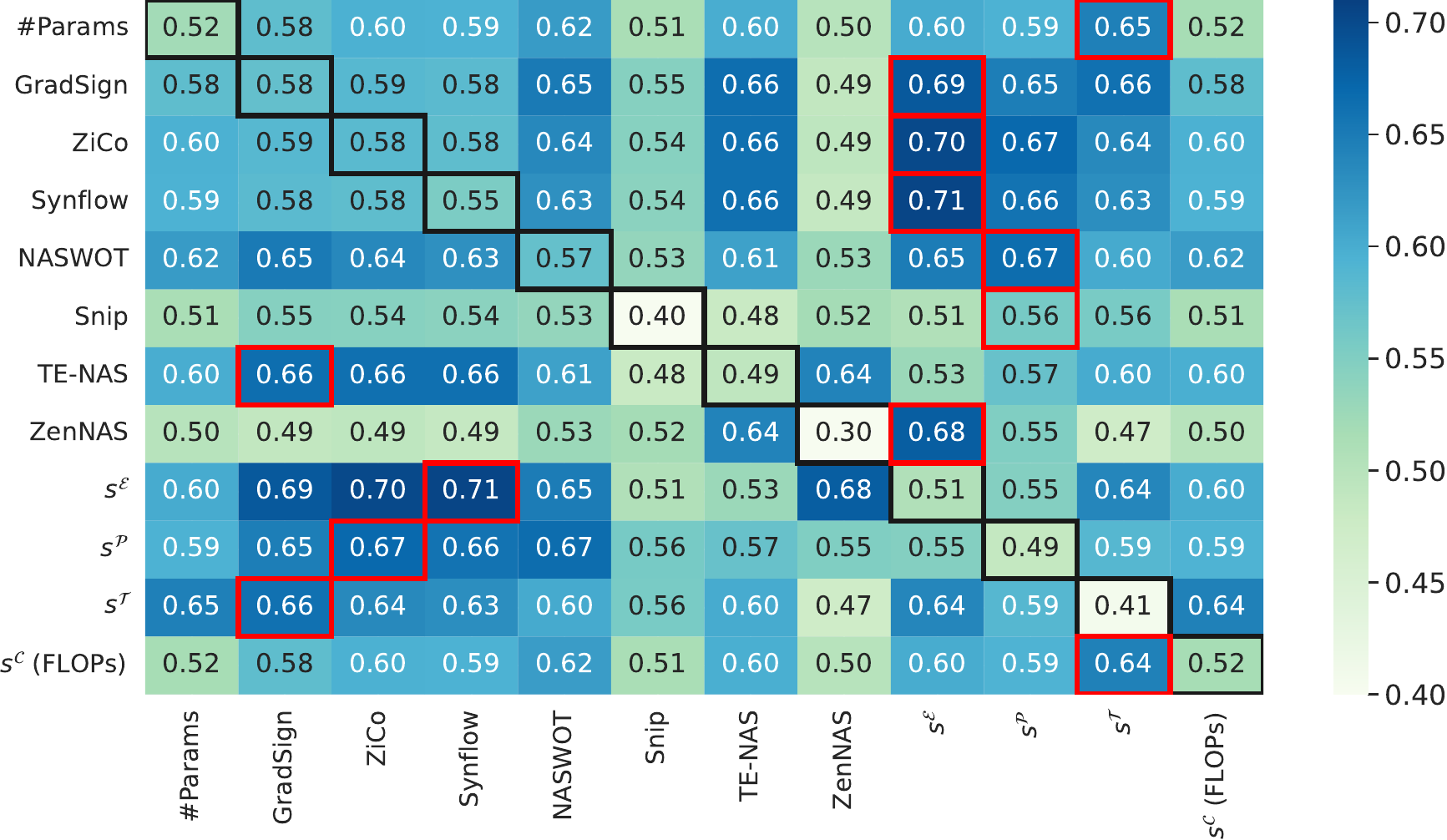}
    \vspace{-0.1cm}
  \caption{Ranking consistency~(Kendall's~$\tau$) w.r.t the performance on ImageNet16-120 of NAS-Bench-201~\cite{dong2020nasbench201} obtained with various combinations of zero-cost proxies. We combine two network rankings, predicted by zero-cost proxies in the $x$- and $y$-axes, using the non-linear ranking aggregation method. The result in the diagonal entry~(represented by a black box) is obtained with the single proxy for the corresponding entry. For each proxy in the $y$-axis, we specify the best combination providing the highest Kendall's~$\tau$ by a red box in the corresponding row.}
  \label{fig:improvement}
  \end{center} 
  \vspace{-0.4cm}
\end{figure*}

\vspace{-0.25cm}
\paragraph{Intercorrelation and complementary features.}
AZ-NAS is built upon the idea that assembling multiple zero-cost proxies can significantly boost the NAS performance. However, naively assembling previous zero-cost proxies might be less effective, mainly due to a lack of complementary features. To investigate this, we analyze in Fig.~\ref{fig:intercorrelation} intercorrelation between network rankings estimated by zero-cost proxies on ImageNet16-120 of NAS-Bench-201~\cite{dong2020nasbench201}. We also compare in Fig.~\ref{fig:improvement} the ranking consistency w.r.t the performance on ImageNet16-120 of NAS-Bench-201, where network rankings predicted by two of the proxies in the $x$ and $y$-axes are integrated.

From these results, we have the following observations: (1)~We can see in Fig.~\ref{fig:intercorrelation} that many existing zero-cost proxies, especially the ones exploiting gradients~(\eg, GradSign~\cite{zhang2022gradsign}, ZiCo~\cite{li2023zico}, and Synflow~\cite{tanaka2020pruning}), predict similar network rankings that are strongly correlated to each other. This coincides with the finding that several gradient-based zero-cost proxies are related theoretically~\cite{shu2022unifying}, providing analogous NAS results. This suggests that these proxies lack complementary features, and thus combining them hardly improves the NAS performance as shown in Fig.~\ref{fig:improvement}. (2)~A few zero-cost proxies~(\eg, GradSign and ZiCo) show strong correlations with \#Params. In their implementation, they compute a proxy score for each weight parameter (or trainable layer), and then obtain the final score of a network by adding all the scores over the weight parameters (or the trainable layers). This implies that they inherently prefer networks with large \#Params, since the final scores are likely to become higher when a larger \#Params are used, resulting in a limited synergy effect among these proxies and \#Params as shown in Fig.~\ref{fig:improvement}. (3)~In terms of the NAS performance using a single proxy only, ZenNAS~\cite{lin2021zen} and Snip~\cite{lee2019snip} show weak ranking consistency w.r.t the performance, \ie, Kendall's~$\tau$ of 0.30 and 0.40, respectively. We can see in Fig.~\ref{fig:improvement} that they rather degrade the NAS performance of several zero-cost proxies~(\eg, \#Params, GradSign, ZiCo, Synflow, or NASWOT~\cite{mellor2021neural}) after integrating the network rankings, suggesting that they are less suitable for an ensemble. On the contrary, while our trainability proxy~($s^{\mathcal{T}}$) also exhibits a weak correlation with the performance, it can help other proxies to improve the ranking consistency by large margins. This clearly demonstrates that the trainability proxy captures a useful network trait for NAS, which is often overlooked by other zero-cost proxies, despite its weak ranking consistency w.r.t the performance. (4)~We can see in Fig.~\ref{fig:intercorrelation} that our zero-cost proxies are less correlated with most of the existing ones. We can also observe in Fig.~\ref{fig:improvement} that incorporating our proxies into others boosts the ranking consistency drastically. This suggests that our proxies offer distinct and useful cues for training-free NAS that could not be identified by other proxies, highlighting their complementary features. In particular, the proxies leveraging gradients~(\eg, GradSign, ZiCo, Synflow, and Snip) typically show the best ensemble results when they are coupled with either the expressivity~($s^{\mathcal{E}}$) or progressivity~($s^{\mathcal{P}}$) proxies that analyze activations. This implies that activations and gradients provide different network characteristics useful for training-free NAS, and it is difficult to achieve good NAS performance when relying solely on one of them. These results confirm once more the importance of the comprehensive evaluation of a network from various and complementary perspectives for effective training-free NAS. (5)~In the case of combining two zero-cost proxies in Fig.~\ref{fig:improvement}, NASWOT and TE-NAS~\cite{chen2020tenas} could be good alternatives for improving the NAS performance of other zero-cost proxies. However, as mentioned in the main paper, NASWOT is only applicable for networks adopting ReLU non-linearities, and TE-NAS is computationally expensive. We can also see that coupling one of the zero-cost proxies of AZ-NAS with \eg, GradSign, ZiCo, or Synflow provides better ranking consistency w.r.t the performance, compared to the combinations among our proxies. Nevertheless, they are outperformed by our AZ-NAS method that assembles all of our zero-cost proxies, which offers a good balance between efficiency and the NAS performance without additional complexities~(see Sec.~4.2 in the main paper for details).

\begin{table}[t]
  \small
  \setlength{\tabcolsep}{0.15cm}
  \centering
  \caption{Search objectives with varying target values of the trainability score. Note that the trainability score ranges from $-\infty$ to $0$. We also report the top-1 validation accuracies on ImageNet~\cite{deng2009imagenet} for the networks found with corresponding configurations.}
  \vspace{-0.3cm}
  \begin{adjustbox}{max width=\columnwidth}
     \begin{tabular}{l c c c c c}
        \toprule
        Name & $s^{\mathcal{E}}$ & $s^{\mathcal{P}}$ & $s^{\mathcal{T}}$ & $s^{\mathcal{C}}$ & Top-1 acc. \\
        \midrule
        \multirow{2}{*}{Model\#1} & \multirow{2}{*}{Maximize} & \multirow{2}{*}{Maximize} & \multirow{2}{*}{\shortstack{Maximize\\(Close to $0$)}}        & \multirow{2}{*}{Maximize} & \multirow{2}{*}{76.55} \\ \\
        Model\#2 & Maximize & Maximize & Close to $-0.2$ & Maximize & 75.12 \\
        Model\#3 & Maximize & Maximize & Close to $-0.3$ & Maximize & 74.37 \\
        Model\#4 & Maximize & Maximize & Close to $-0.4$ & Maximize & 72.68 \\
        Model\#5 & Maximize & Maximize & Close to $-0.5$ & Maximize & 71.10 \\
        \bottomrule
     \end{tabular} \label{tab:varying_trainability}
  \end{adjustbox}
  \vspace{-0.2cm}
\end{table}

\begin{figure}[t]
  \captionsetup[subfigure]{justification=centering}
  \begin{center}
    \begin{subfigure}[t]{0.495\columnwidth}
      \centering
      \includegraphics[keepaspectratio,height=3.3cm]{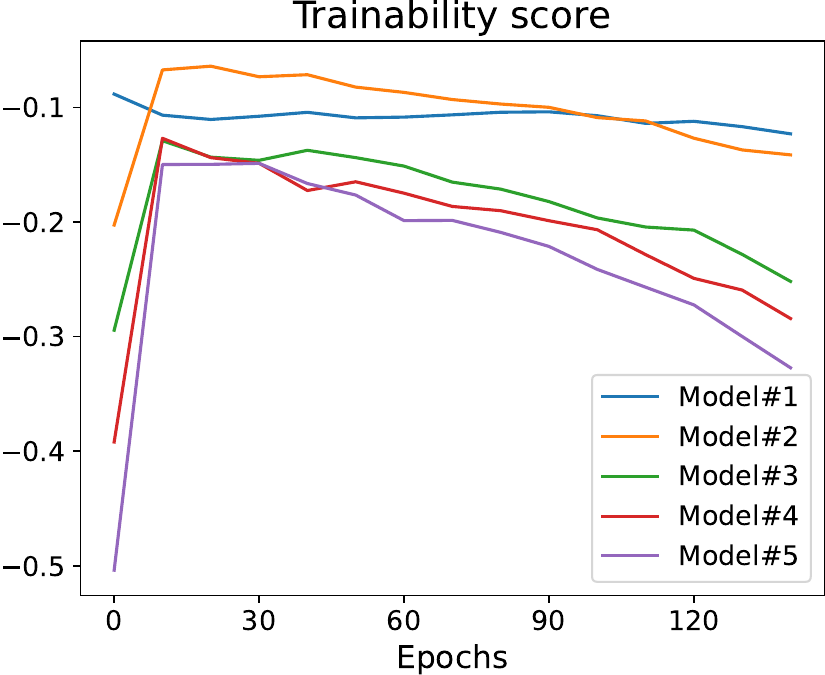}
      \caption{Trainability scores~$s^{\mathcal{T}}$ \\~~~for every 10 epochs.}
    \end{subfigure}
    \begin{subfigure}[t]{0.495\columnwidth}
      \centering
      \includegraphics[keepaspectratio,height=3.3cm]{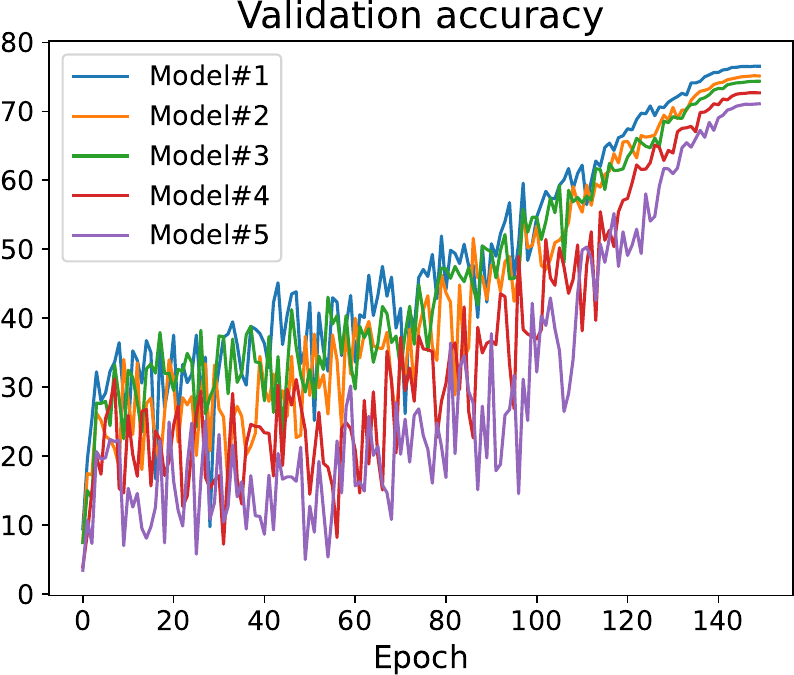}
      \caption{Top-1 validation accuracy.}
    \end{subfigure}

    \vspace{0.2cm}

    \begin{subfigure}[t]{0.495\columnwidth}
      \centering
      \includegraphics[keepaspectratio,height=3.3cm]{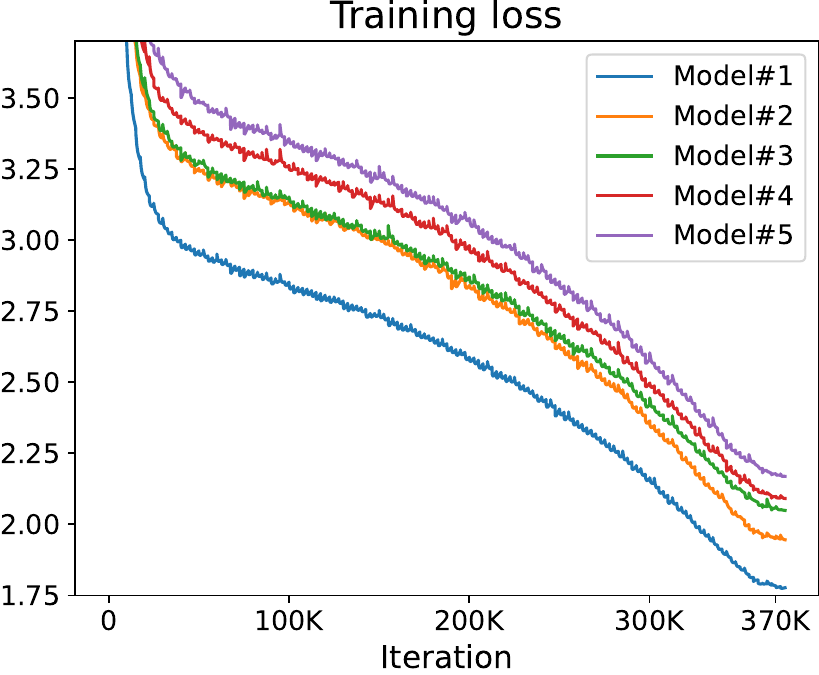}
      \caption{Training loss.}
    \end{subfigure}
    \begin{subfigure}[t]{0.495\columnwidth}
      \centering
      \includegraphics[keepaspectratio,height=3.3cm]{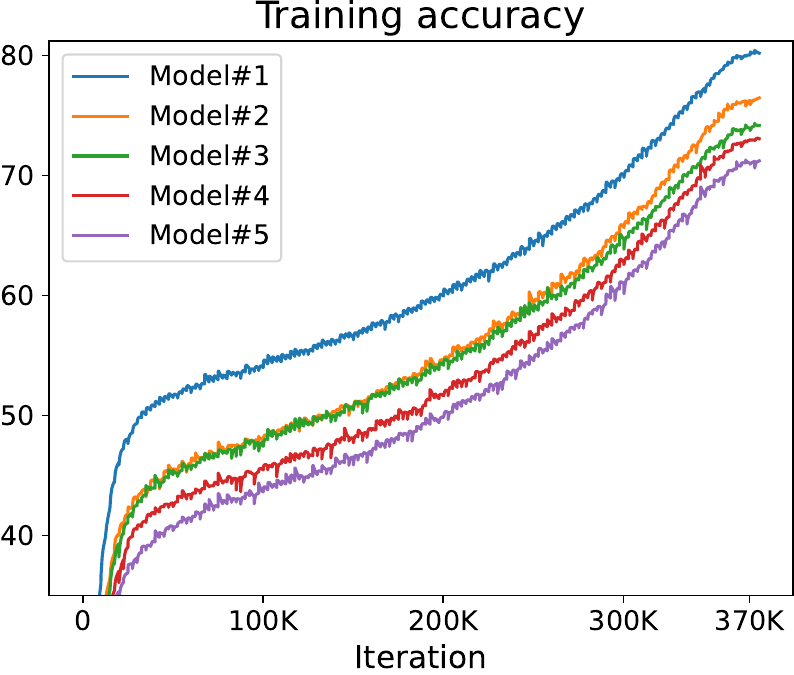}
      \caption{Training accuracy.}
    \end{subfigure}

    \vspace{0.2cm}

    \begin{subfigure}[t]{0.495\columnwidth}
      \centering
      \includegraphics[keepaspectratio,height=3.3cm]{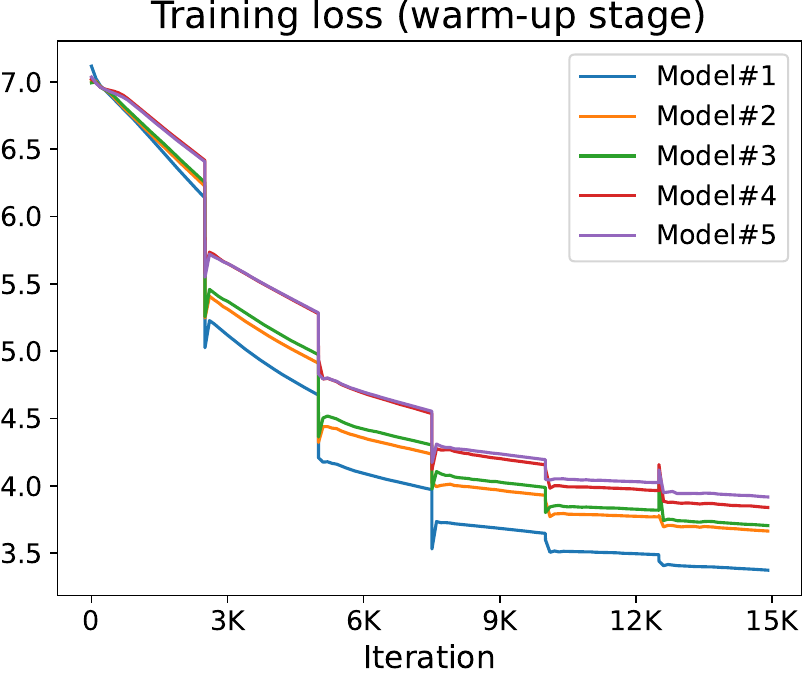}
      \caption{Training loss \\~~~(warm-up stage).}
    \end{subfigure}
    \begin{subfigure}[t]{0.495\columnwidth}
      \centering
      \includegraphics[keepaspectratio,height=3.3cm]{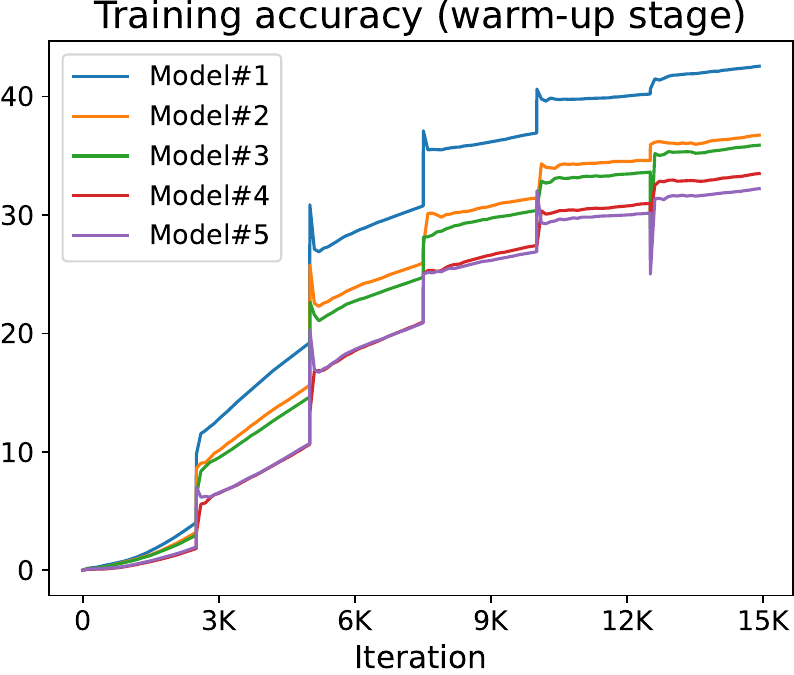}
      \caption{Training accuracy \\~~~(warm-up stage).}
    \end{subfigure}
  \end{center}
        \vspace{-0.5cm}
        \caption{Comparison of the networks with different trainability scores, where they are found with the search objectives specified in Table~\ref{tab:varying_trainability}. We show in (a) and (b) the trainability scores of the networks and validation accuracy on ImageNet~\cite{deng2009imagenet} along epochs, respectively. We also compare in (c)-(f) the training curves of the networks. (Best viewed in color.)} \label{fig:varying_trainability}
        \vspace{-0.3cm}
\end{figure}

\renewcommand\fbox{\fcolorbox{white}{white}}
\begin{figure*}[t]
  \small
  \begin{center}
    \begin{subfigure}{0.29\textwidth}  
      \fbox{\parbox[c][2cm]{0.95\textwidth}{
          \centering
          \raisebox{0cm}{\includegraphics[scale=0.285]{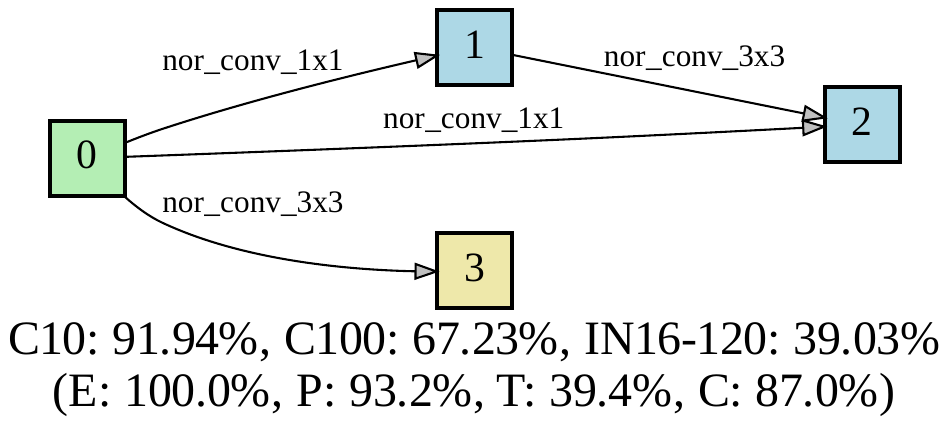}}
      }}
      \fbox{\parbox[c][1.8cm]{0.95\textwidth}{
          \centering
          \raisebox{-1.8cm}{\includegraphics[scale=0.285]{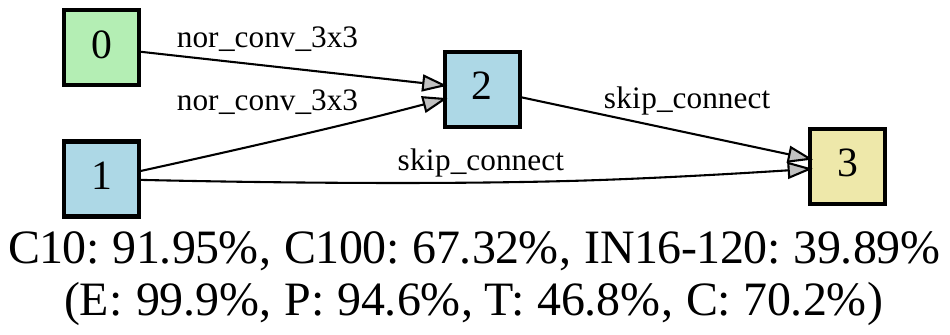}}
      }}
      \fbox{\parbox[c][2cm]{0.95\textwidth}{
          \centering
          \raisebox{-2.05cm}{\includegraphics[scale=0.285]{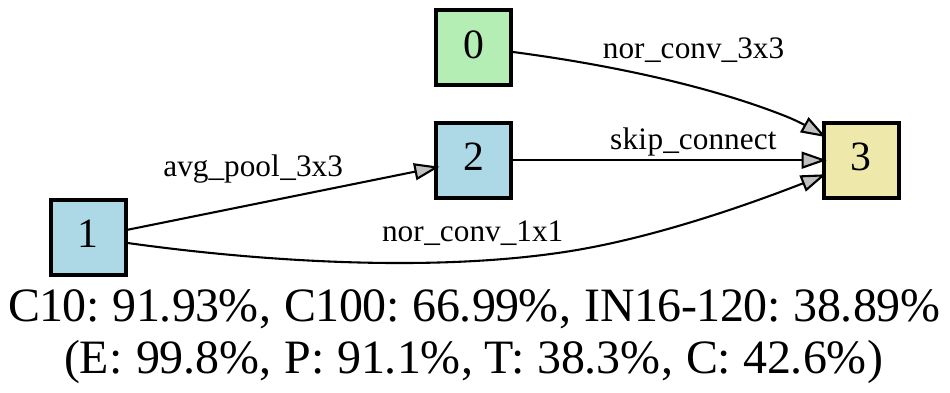}}
      }}
      \caption{Top-3 architectures selected with~$s^{\mathcal{E}}$.}
    \end{subfigure}
    \hspace{.15cm}
    \begin{subfigure}{0.29\textwidth}  
      \fbox{\parbox[c][2cm]{0.95\textwidth}{
          \centering
          \raisebox{-2.05cm}{\includegraphics[scale=0.285]{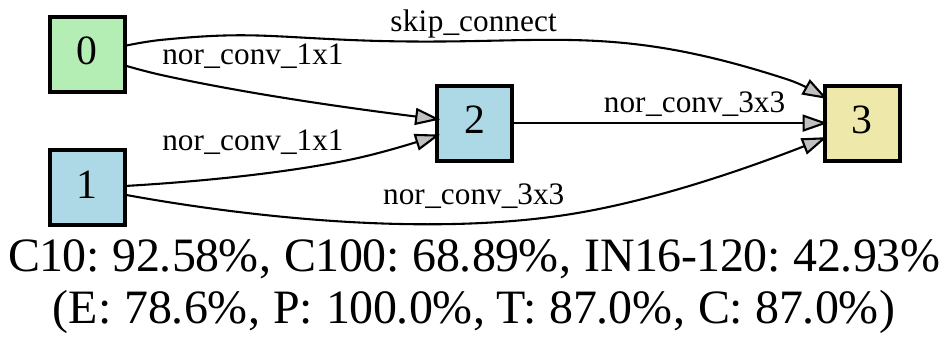}}
      }}
      \fbox{\parbox[c][1.8cm]{0.95\textwidth}{
          \centering
          \raisebox{-1.81cm}{\includegraphics[scale=0.285]{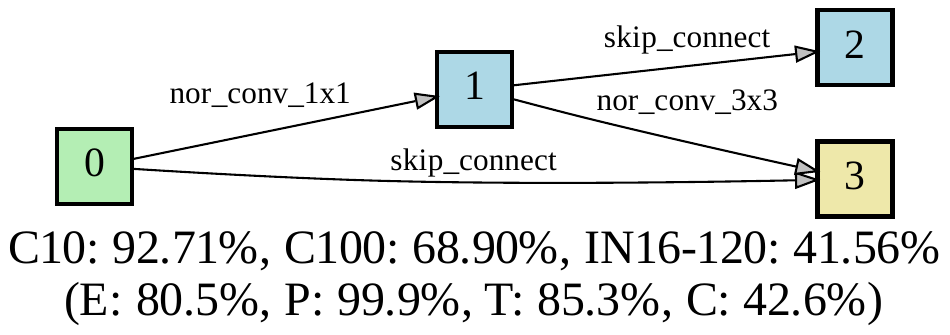}}
      }}
      \fbox{\parbox[c][2cm]{0.95\textwidth}{
          \centering
          \raisebox{0cm}{\includegraphics[scale=0.285]{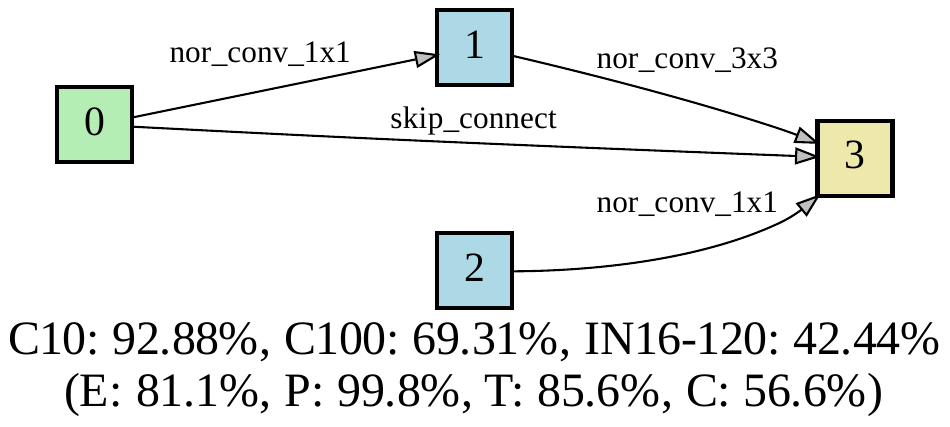}}
      }}
      \caption{Top-3 architectures selected with~$s^{\mathcal{P}}$.}
    \end{subfigure}
    \hspace{.15cm}
    \begin{subfigure}{0.385\textwidth}
      \fbox{\parbox[c][2cm]{0.95\textwidth}{
          \centering
          \raisebox{-2.05cm}{\includegraphics[scale=0.285]{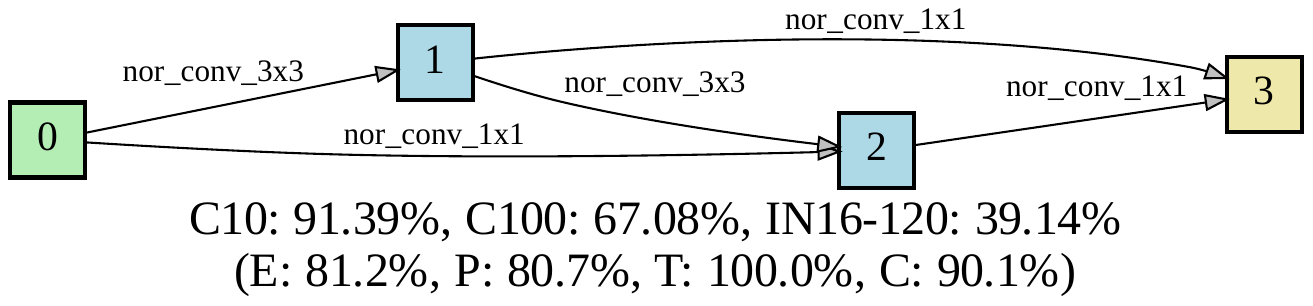}}
      }}
      \fbox{\parbox[c][1.8cm]{0.95\textwidth}{
          \centering
          \raisebox{0cm}{\includegraphics[scale=0.285]{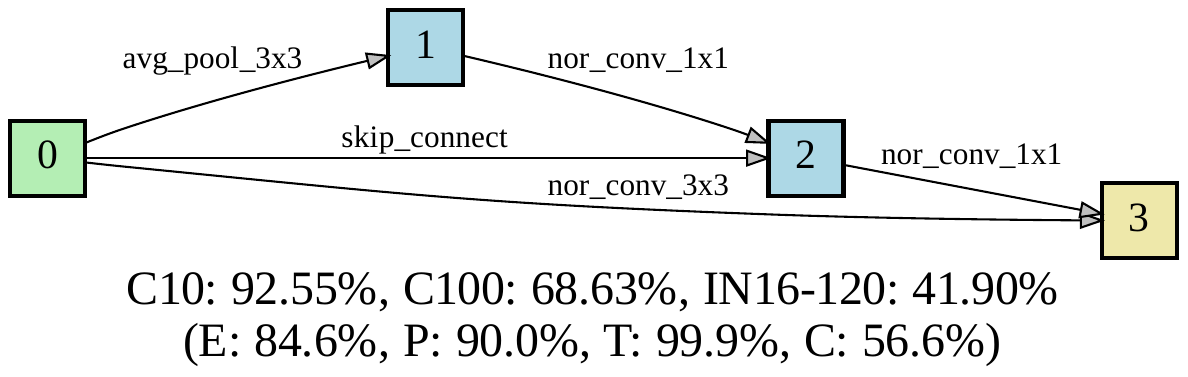}}
      }}
      \fbox{\parbox[c][2cm]{0.95\textwidth}{
          \centering
          \raisebox{-2.05cm}{\includegraphics[scale=0.285]{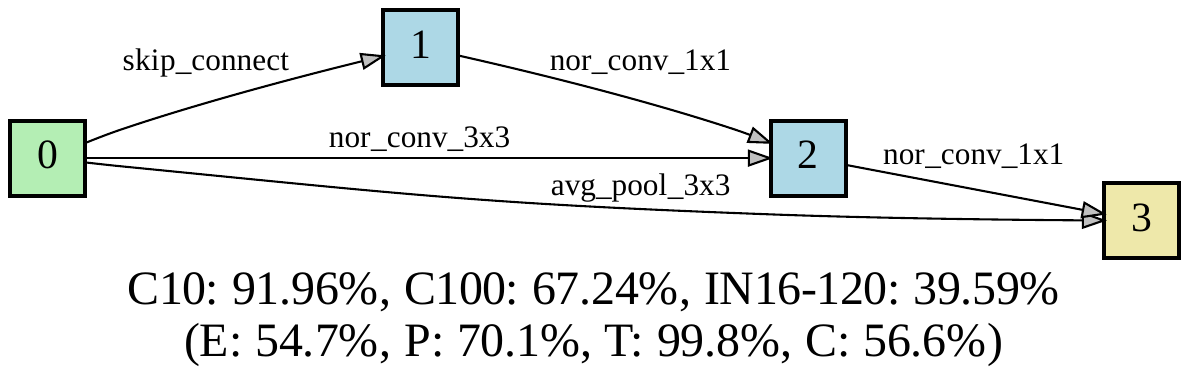}}
      }}
      \caption{Top-3 architectures selected with~$s^{\mathcal{T}}$.}
    \end{subfigure}

    \vspace{0.6cm}

    \begin{subfigure}{0.385\textwidth}
      \fbox{\parbox[c][1.75cm]{0.95\textwidth}{
          \centering
          \raisebox{-1.82cm}{\includegraphics[scale=0.285]{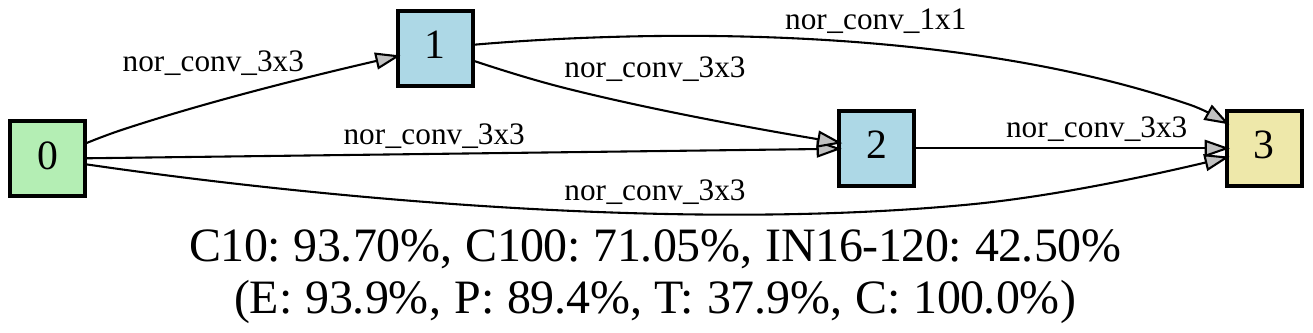}}
      }}
      \fbox{\parbox[c][1.55cm]{0.95\textwidth}{
          \centering
          \raisebox{0cm}{\includegraphics[scale=0.285]{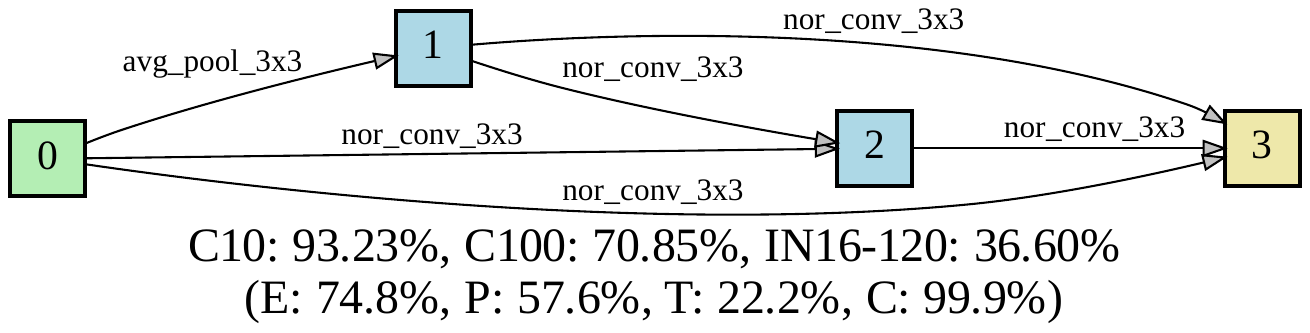}}
      }}
      \fbox{\parbox[c][1.55cm]{0.95\textwidth}{
          \centering
          \raisebox{0cm}{\includegraphics[scale=0.285]{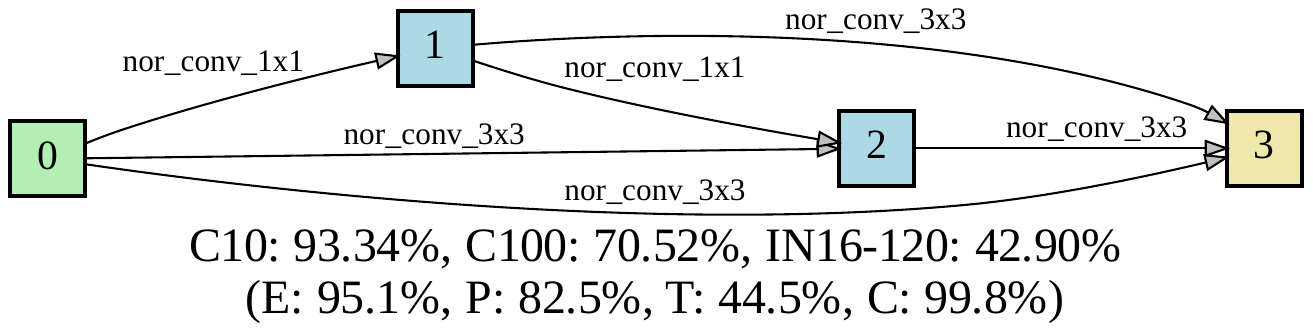}}
      }}
      \caption{Top-3 architectures selected with~$s^{\mathcal{C}}$.}
    \end{subfigure}
    \hspace{1.8cm}
    \begin{subfigure}{0.385\textwidth}
      \fbox{\parbox[c][1.75cm]{0.95\textwidth}{
          \centering
          \raisebox{0cm}{\includegraphics[scale=0.285]{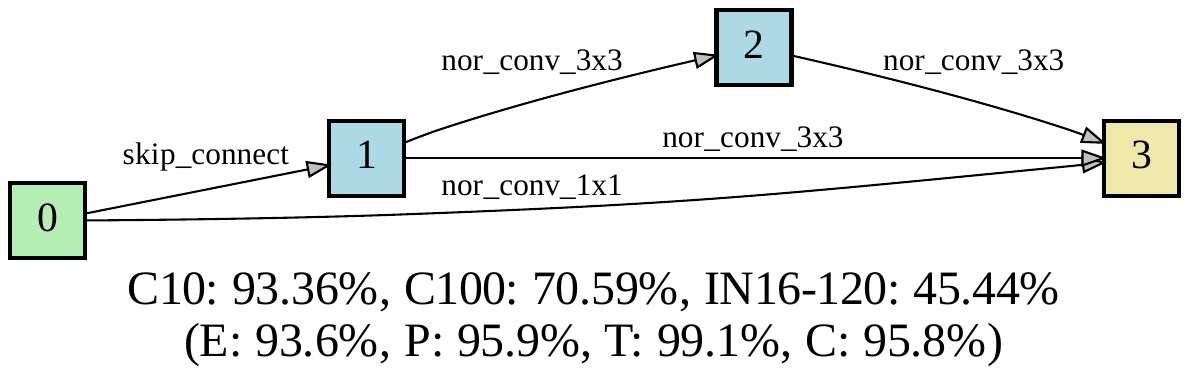}}
      }}
      \fbox{\parbox[c][1.55cm]{0.95\textwidth}{
          \centering
          \raisebox{0cm}{\includegraphics[scale=0.285]{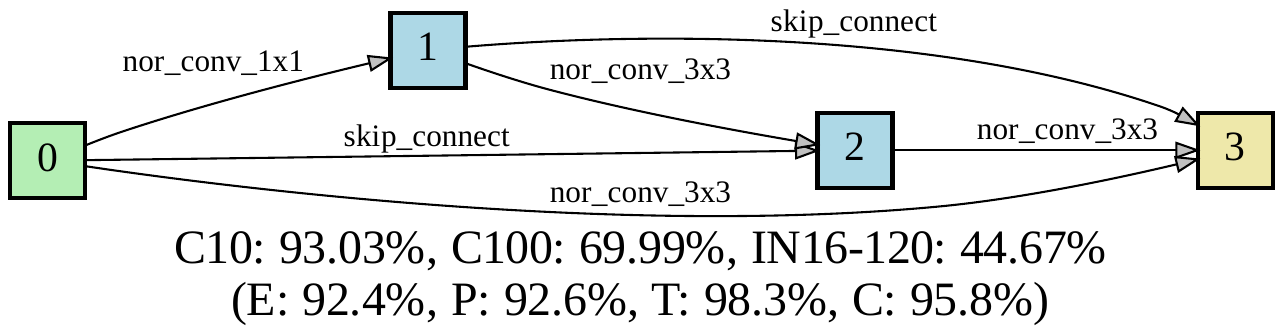}}
      }}
      \fbox{\parbox[c][1.55cm]{0.95\textwidth}{
          \centering
          \raisebox{0cm}{\includegraphics[scale=0.285]{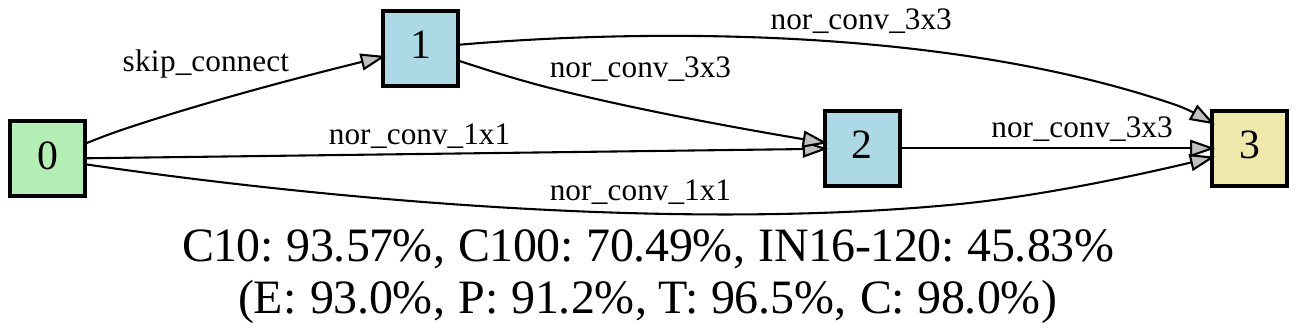}}
      }}
      \caption{Top-3 architectures selected with~$s^{\textrm{AZ}}$.}
    \end{subfigure}
  \end{center}
  \vspace{-0.3cm}
     \caption{Visualization of the top-3 network architectures found with each zero-cost proxy score of AZ-NAS and the final AZ-NAS score on NAS-Bench-201~\cite{dong2020nasbench201}. The green and yellow squares indicate the input and output nodes of a cell, respectively, and the blue ones represent intermediate nodes. For each architecture, we report the top-1 test accuracies on the CIFAR-10/100~(C10/100) and ImageNet16-120~(IN16-120) datasets, together with the percentiles for the proxy scores in parenthesis, where 100\% indicates the highest ranking. We denote by E, P, T, and C the expressivity, progressivity, trainability, and complexity proxies, respectively, for brevity.}
  \label{fig:vis_archs}
  \vspace{-0.2cm}
\end{figure*}

\vspace{-0.25cm}
\paragraph{Analysis on the trainability proxy.}
The zero-cost proxies of AZ-NAS assess a network with randomly initialized weight parameters. The expressivity and progressivity proxies could benefit from this setting, since they can evaluate the feature space of a network by analyzing diverse features. That is, the features extracted by a randomly initialized network could occupy the space to its maximum capacity as they are randomly distributed along various orientations. The complexity proxy measures FLOPs of a network architecture, which remains fixed regardless of training. Similarly, the trainability proxy evaluates a network at initialization in terms of stable gradient propagation, focusing on the spectral norm of the Jacobian matrix for a primary block. Note that the stable gradient propagation of a network at the initial state has been proven to be crucial for high performance~\cite{glorot2010understanding,he2015delving,qi2023lipsformer}, supporting our idea of scoring the trainability proxy without training. 

To further demonstrate the validity of our approach to measuring the trainability score at initialization, we perform an in-depth analysis of the trainability proxy on the MobileNetV2 search space~\cite{sandler2018mobilenetv2,lin2021zen}. Specifically, we select five distinct architectures by setting different search objectives of the trainability proxy for the evolutionary algorithm, and train them with the same training scheme used for the experiments in Table~\ref{tab:MBV2_reproduce}. We summarize the search configurations and results in Table~\ref{tab:varying_trainability}. From the table, we can observe that the final performance of a network is largely affected by the trainability score, where a network with a higher trainability score shows better performance. We also present in Fig.~\ref{fig:varying_trainability} how the trainability score changes during training, and its impact on training and performance, based on the networks chosen in Table~\ref{tab:varying_trainability}. We can see from Fig.~\ref{fig:varying_trainability}(a) that the trainability scores are not maintained during training, possibly because they are affected by weight parameters that keep changing during training. Nevertheless, the relative ranking between them is roughly preserved along training epochs. This suggests that a high trainability score at initialization is important for a network to achieve better performance, as evidenced by Fig.~\ref{fig:varying_trainability}(b). We can also see from Figs.~\ref{fig:varying_trainability}(c) and (d) that the better trainability score consistently results in better training losses and accuracies throughout training iterations. In particular, we can find in Figs.~\ref{fig:varying_trainability}(e) and (f) that the rankings of training losses and accuracies are aligned with the ranking of the trainability scores even at the very beginning of training~(\ie,~the warm-up stage), highlighting the importance of the trainability proxy at the initial state of a network. This finding also coincides with the concept of learning curve extrapolation~\cite{domhan2015speeding} or an early stopping technique for NAS~\cite{baker2017accelerating}. These results confirm that considering our trainability proxy at initialization is effective for training-free NAS, and it plays a significant role in predicting the ranking of candidate networks in terms of the final performance.

\vspace{-0.25cm}
\paragraph{Visualization of selected architectures.}
We visualize in Fig.~\ref{fig:vis_archs} network architectures chosen by individual zero-cost proxies of AZ-NAS or the final AZ-NAS score on NAS-Bench-201~\cite{dong2020nasbench201}. We select top-3 network architectures for each proxy among 1042 candidate networks, where the candidates are sampled evenly according to the test accuracy on ImageNet16-120 of NAS-Bench-201. For each architecture, we report the test accuracies on CIFAR-10/100 and ImageNet16-120, as well as the percentiles of our zero-cost proxy scores~(\ie, 100\% indicates that a network exhibits the highest score). We can see from Figs.~\ref{fig:vis_archs}(a)-(d) that relying solely on one of the zero-cost proxies causes structural biases in the selected architectures. For example, the expressivity proxy in Fig.~\ref{fig:vis_archs}(a) consistently assigns high scores to networks with a cell structure consisting of a single convolution layer with a kernel size of $3 \times 3$, whereas the progressivity proxy in Fig.~\ref{fig:vis_archs}(b) favors networks stacking convolutional layers with kernel sizes of~$1 \times 1$ and~$3 \times 3$ with an additional skip connection. These proxies also introduce unnecessary intermediate nodes within a cell structure, whose input or output edges are disconnected. The complexity proxy in Fig.~\ref{fig:vis_archs}(d) tends to prefer networks in which all the edges of a cell structure are defined as parametric operations. Such biases prevent us from finding high-performing networks, degrading the NAS performance. We can also observe that the networks chosen by a single zero-cost proxy exhibit low scores for the other proxies frequently, leading to unsatisfactory test accuracies, especially on ImageNet16-120 that includes images depicting more complex scenes and objects. This highlights that exploiting a single proxy solely is insufficient for evaluating a network without training. In contrast, assembling the zero-cost proxies in Fig.~\ref{fig:vis_archs}(e) allows us to choose networks highly-ranked for all the proxies, which show high performance across the datasets, without suffering from a specific structural bias.

\end{document}